\documentclass{article} % For LaTeX2e
\usepackage{iclr2026_conference,times}

\usepackage{amsmath,amsfonts,bm}

\def\eqref#1{equation~\ref{#1}}
\def\1{\bm{1}}

\DeclareMathAlphabet{\mathsfit}{\encodingdefault}{\sfdefault}{m}{sl}
\SetMathAlphabet{\mathsfit}{bold}{\encodingdefault}{\sfdefault}{bx}{n}

\usepackage{subcaption}
\usepackage{adjustbox}
\usepackage{booktabs}
\usepackage{pdfpages}
\usepackage{graphicx}
\usepackage{multirow}
\usepackage{wrapfig}
\usepackage{pgfplots}
\usepackage{caption}
\usepackage{marvosym}
\usepackage{float}
\usepackage{hyperref}
\hypersetup{
    pageanchor=false
}

\title{Explore-on-Graph: Incentivizing Autonomous Exploration of Large Language Models on Knowledge Graphs with Path-refined Reward Modeling}
\author{
\begin{tabular}{l} % "l" 表示左对齐
    \\
    \textbf{Shiqi Yan$^{1,2}$, Yubo Chen$^{1}$, Ruiqi Zhou$^{2}$, Zhengxi Yao$^{2}$, Shuai Chen$^{4}$, Tianyi Zhang$^{4}$,} \\
    \textbf{Shijie Zhang$^{4}$, WeiQiang Zhang$^{1,2}$, Yongfeng Huang$^{1,2}$, Haixin Duan$^{1,3}$, Yunqi Zhang$^{1}$\thanks{Corresponding author. Email: \texttt{zhangyq@mail.zgclab.edu.cn}}} \\
    \textnormal{$^1$Zhongguancun Laboratory, Beijing, China} \\
    \textnormal{$^2$Department of Electronic Engineering, Tsinghua University, Beijing, China} \\
    \textnormal{$^3$Institute for Network Sciences and Cyberspace, Tsinghua University, Beijing, China} \\
    \textnormal{$^4$Ant International, Ant Group, Hangzhou, Zhejiang, China}
  \end{tabular}
}
\iclrfinalcopy % Uncomment for camera-ready version, but NOT for submission.
\begin{document}

\maketitle

% \begin{abstract}
% Reasoning over knowledge graphs is a fundamental challenge in artificial intelligence, requiring both accurate inference and interpretability. Recent advances leverage reinforcement learning to optimize reasoning paths, yet most approaches focus on end-state rewards, neglecting the importance of intermediate reasoning steps. In this work, we propose a deliberative and constructive reasoning framework that explicitly rewards the process of path exploration on knowledge graphs. By integrating process-based rewards into policy optimization, our method encourages agents to discover interpretable and effective reasoning trajectories. Experiments on benchmark datasets demonstrate improved performance in both answer accuracy and path quality, highlighting the significance of path-aware policy design for knowledge graph reasoning.

% \end{abstract}

\begin{abstract}% .
The reasoning process of  Large Language Models (LLMs) is often plagued by hallucinations and missing facts in question-answering tasks.
A promising solution is to ground LLMs' answers in verifiable knowledge sources, such as Knowledge Graphs (KGs).
Prevailing KG-enhanced methods typically constrained LLM reasoning either by enforcing rules during generation or by imitating paths from a fixed set of demonstrations.
% However, these approaches limits the model's flexibility and exploration capabilities, leading to insufficient utilization of graphs and thus impairing the model's generalization in complex reasoning scenarios.
% However, these approaches restricted LLMs to previously seen patterns, fundamentally limiting their generalization to complex graphs with unseen reasoning structures.
% this reliance on static and pre-defined patterns fundamentally limit the model's ability to explore unseen paths of the graph, thus impairing its generalization to more complex and novel reasoning paths.
% However, these approaches restrict LLMs to previously seen patterns, fundamentally limiting their generalization to complex graphs. 
However, they naturally confined the reasoning patterns of LLMs within the scope of prior experience or fine-tuning data, limiting their generalizability to out-of-distribution graph reasoning problems.
% To tackle this problem,
To address this issue,
% , we aim to enhance LLMs' reasoning capability by encouraging to explore a more diverse reasoning space.
% To overcome this, our goal is exposing the model to a broader and richer reasoning patterns
% is to promote the generalization through autonomous exploration, exposing the model to a broader and more complex set of reasoning patterns.
% they usually suffered from the inherent inflexibility of these prescribed paradigms,
% hindering their ability to actively explore the graph for alternative and potentially more effective reasoning process.
% restricts the model to the demonstrated patterns and discourages it from discovering alternative, potentially more efficient or novel, reasoning strategies.
in this paper, we propose Explore-on-Graph (EoG), a novel framework that encourages LLMs to autonomously explore a more diverse reasoning space on KGs.
To incentivize exploration and discovery of novel reasoning paths, we propose to introduce reinforcement learning during training, whose reward is the correctness of the reasoning paths' final answers. 
To enhance the efficiency and meaningfulness of the exploration, we propose to incorporate path information as additional reward signals to refine the exploration process and reduce futile efforts.
Extensive experiments on five KGQA benchmark datasets demonstrate that, to the best of our knowledge, our method achieves state-of-the-art performance, outperforming not only open-source but also even closed-source LLMs\footnote{Code and data are available at: https://github.com/ysq111333/EoG}. 
\end{abstract}

\section{Introduction}
% Large Language Models (LLMs) have demonstrated remarkable capabilities 
% in various natural language processing tasks\citep{brown2020language, chowdhery2022palm, touvron2023llama}.

% Large Language Models (LLMs) have achieved remarkable success in various 
% natural language processing tasks\citep{brown2020language, chowdhery2022palm, touvron2023llama},
% demonstrating strong reasoning abilities 
% in complex scenarios\citep{wei2022chain, kojima2023large, zelikman2022star}.
% Large Language Models (LLMs) have achieved demonstrated remarkable reasoning abilities
% in various natural language processing tasks\citep{brown2020language, }
% Despite their impressive capabilities, LLMs often struggle with hallucinations and
% factual inaccuracies\citep{ji2023survey, zhang2023survey}, particularly
% when answering questions that require specific knowledge.
% To address these challenges, 
% Knowledge Graph Question Answering (KGQA) aims to answer natural language questions
% by reasoning over large-scale knowledge graphs (KGs).

Large language models (LLMs) have demonstrated impressive capabilities
in various natural language processing tasks~\citep{Brown2020gpt3,team2023gemini,liu2024deepseek,dubey2024llama}.
Despite their success, LLMs often struggle to produce faithful reasoning in question-answering (QA) tasks~\citep{wang2023knowledge,radhakrishnan2023question}, primarily due to knowledge gaps and
their susceptibility to hallucination~\citep{ji2023survey,guan2024mitigating}. 
To this end, the reliability of LLMs in real-world QA systems remains a significant concern.

To mitigate these issues, a promising approach is to ground LLMs' responses in external and verifiable knowledge sources~\citep{lewis2020retrieval,li2023few,ren2025investigating}.
Knowledge graphs (KGs), which emerge as an ideal candidate, provide solid factual information to validate and guide the LLMs' reasoning process~\citep{luo2024rog,zhu2024llms}. 

% Knowledge Graph Question Answering (KGQA) is defined as automatically answering
% natural language questions by reasoning over the relevant relational triples in large-scale knowledge graphs.
% It is a critical task for natural language understanding, especially for
% improving the factuality and mitigating the hallucinations of large language models (LLMs)
% through grounding their reasoning on verifiable and structured facts.
% Consequently, the potential of knowledge graph question answering (KGQA) has spurred increasing research interest.
Recent KG-enhanced reasoning methods can be broadly classified into two paradigms:
rule-based methods and imitation-based methods. 
The former~\citep{sun2023thinkongraph,zhang-etal-2025-rule} typically leveraged a pre-defined set of rules to ensure the logical consistency and faithfulness during the LLM's reasoning process.
% These methods typically leverage a pre-defined set of logical rules to prune the search space at each decoding step.
The latter~\citep{zhang2022subgraph,wu2024cotkr,jiang-etal-2025-kg} mainly trained LLMs on extensive datasets to directly mimic the reasoning patterns and heuristics embedded in the demonstrations.

% The former involves directly extracting a series of relational triples which serve as candidate reasoning paths, while the latter employs an agent that constructs reasoning paths through an iterative process of querying the graphs.

% while the intention behind schema-guided KG-enhancement is to ground LLMs in factual knowledge and provide a structured reasoning framework, an over-reliance on this approach can inadvertently create a model that is more of a puppet than a partner.
% Existing methods achieved considerable success in constraining the LLM's reasoning process by known features and patterns. 
\begin{figure}[t]
    \centering
    \includegraphics[width=.95\textwidth]{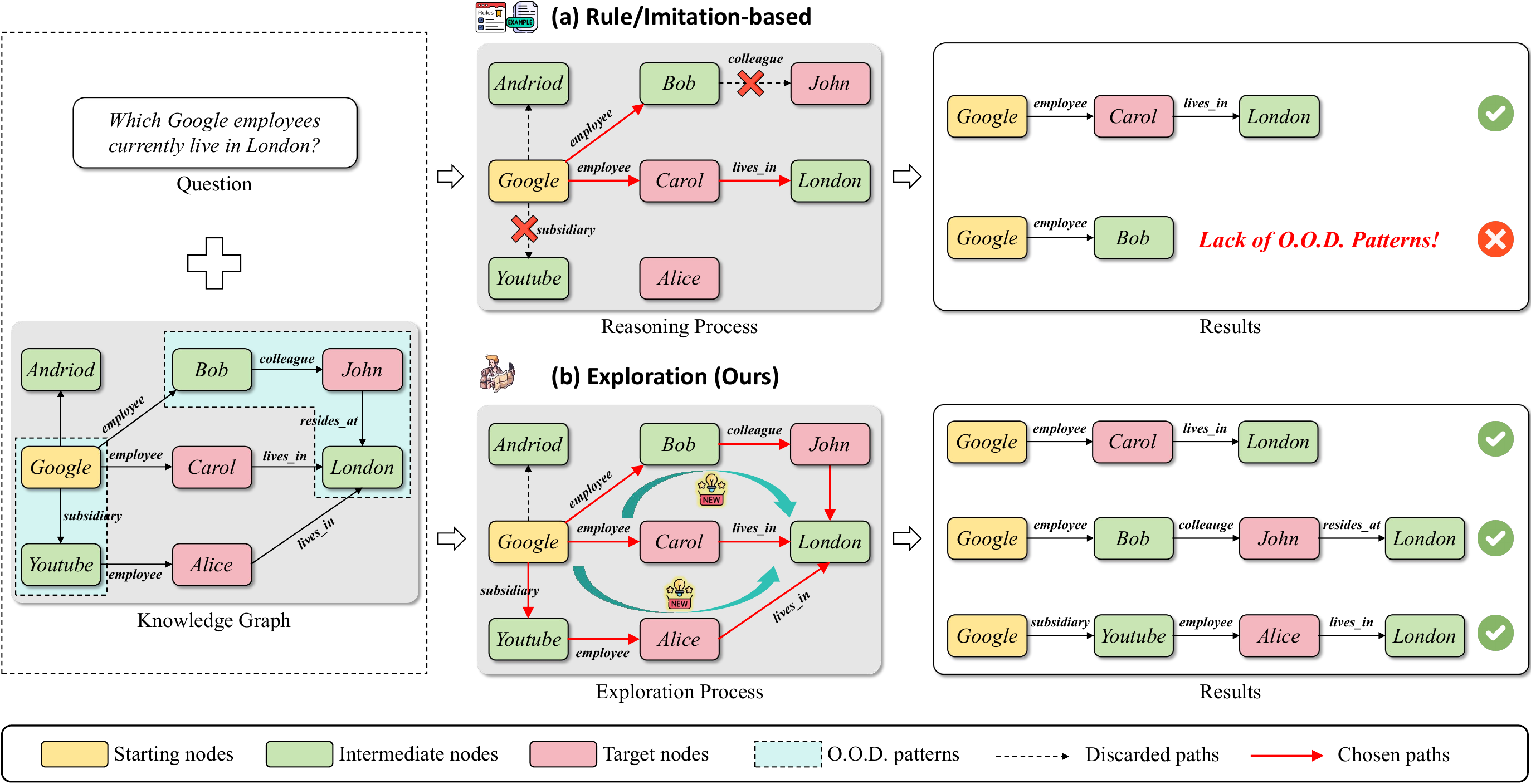}
    \caption{Examples of (a) rule/imitation-based method and (b) our exploration method. O.O.D. stands for Out-Of-Distribution.}
    \label{fig:intro}
\end{figure}

However, existing methods usually failed to generalize to complex graphs, especially to novel reasoning patterns that fall outside the distribution of the fine-tuning data or pre-defined rules.
For example, as shown in Figure \ref{fig:intro} (a), ``$\textit{Google}\xrightarrow{\textit{employee}}\textit{Carol}\xrightarrow{\textit{lives\_in}}\textit{London}$'' is the most common reasoning pattern corresponding to the question, and thus can be successfully recognized by existing methods. However, the patterns ``$\textit{Google}\xrightarrow{\textit{employee}}\textit{Bob}\xrightarrow{\textit{colleague}}\textit{John}\xrightarrow{\textit{resides\_at}}\textit{London}$'' and ``$\textit{Google}\xrightarrow{\textit{subsidiary}}\textit{Youtube}\xrightarrow{\textit{employee}}\textit{Alice}\xrightarrow{\textit{lives\_in}}\textit{London}$'' involve additional steps ``$\xrightarrow{\textit{colleague}}$'' and ``$\xrightarrow{\textit{subsidiary}}$'' that deviate from the most common pattern, consequently rendering them out-of-distribution and challenging for existing methods.
This observation motivates us to believe that navigating beyond the training distribution requires the capability to venture into unfamiliar regions of the graph.
For example, in Figure \ref{fig:intro}(b), although the ``$\xrightarrow{\textit{colleague}}$'' and ``$\xrightarrow{\textit{subsidiary}}$'' patterns are out of distribution, actively exploring these unfamiliar regions effectively helps discover these novel paths and give more accurate answers.
We therefore argue that a complementary capability of autonomous exploration is essential to enhance generalization in the graph reasoning task.

In this paper, we propose a novel framework named Explore-on-Graph (EoG) to incentivize autonomous exploration of LLMs on KGs.
Inspired by recent advances in large reasoning models~\citep{guo2025deepseek}, we propose to introduce reinforcement learning during training, which is preceded by a Supervised Fine-Tuning (SFT) step, to stimulate exploration capabilities.
We use the correctness of the reasoning paths' final answers as reward signals, which can be computed programmatically given the golden label.
Moreover, to improve the efficiency and semantic meaningfulness of the exploration process, we propose to incorporate path information as additional reward signals to refine the exploration strategy.
Specifically, we introduce a second dedicated training stage and leverage our proposed path-refined reward.
To evaluate the effectiveness of our approach, we conducted extensive experiments on five KGQA benchmark datasets. 
Experimental results demonstrate that our method achieves state-of-the-art performance, substantially outperforming a range of baseline methods and even surpassing the results of powerful closed-source models.

\section{Related Work}
% Knowledge Graph Question Answering (KGQA) is a critical task for improving factuality and mitigating hallucinations through grounding their reasoning on verifiable and structured facts.

Knowledge graphs, which consist of relational triples that describe real-world entities and their relationships~\citep{kg_acm,chen-etal-2022-learning,zhang2022relu}, 
have been extensively adopted to guide the reasoning process of large language models to improve factuality and reduce hallucinations~\citep{agrawal2024can}.
Recent approaches to KG-enhanced reasoning could be broadly categorized into two paradigms based on the generation of the reasoning process: rule-based methods~\citep{sun2023thinkongraph,zhang-etal-2025-rule,li-etal-2025-decoding-graphs} and imitation-based methods~\citep{wu2024cotkr,mavromatis2025gnn,jiang-etal-2025-kg}.
To ensure the faithfulness of the reasoning process, rule-based methods guide the LLM's reasoning process on the KGs by pre-defined rules.
For example, ToG~\citep{sun2023thinkongraph} introduced intuitive instructions to prompt LLMs to prune entities and relations from subgraphs. 
To improve retrieval efficiency, ReKnoS~\citep{wang2025reasoning} proposed to recognize super-relations that are defined as a set of semantically related relations.
Moreover, DoG~\citep{li-etal-2025-decoding-graphs} adopted graph-aware constrained decoding with structured chains to constrain the decoding process. 
The approaches were generally training-free, but did not enhance the intrinsic reasoning capabilities of the LLMs themselves.

Imitation-based methods focus on emulating reasoning patterns derived from fine-tuning data.
Early work mainly tried to convert questions into executable logical forms (e.g. SPARQL) for knowledge graph retrieval~\citep{lan2020query,das-etal-2021-case,ye-etal-2022-rng}, which heavily relied on the quality of generated queries.
More recent work widely used Chain of Thought (CoT) to enhance LLM reasoning~\citep{wu2024cotkr,zhao2024kg}.
To reduce incorrect thoughts, RoG~\citep{luo2024rog} proposed a planning-retrieval-reasoning framework that grounds plans in KGs. 
PoG~\citep{chen2024plan} further improved the planning process through a reflection mechanism for self-correction.
Though Kg-Agent~\citep{jiang-etal-2025-kg} employed several agents to iteratively reason over KGs, it relied on supervised fine-tuning with synthesized program data and failed to generalize beyond pre-defined tool-based reasoning paths.

Different from previous work, we propose to incentivize autonomous exploration via reinforcement learning with path-refined reward modeling, enabling the model to explore novel reasoning paths that fall outside the distribution of pre-defined rules or supervised fine-tuning data.
Experimental results on several benchmark datasets prove the effectiveness of our method.

\section{Methodology}
\label{headings}
\begin{figure*}[t]
    \centering
    \includegraphics[width=.95\textwidth]{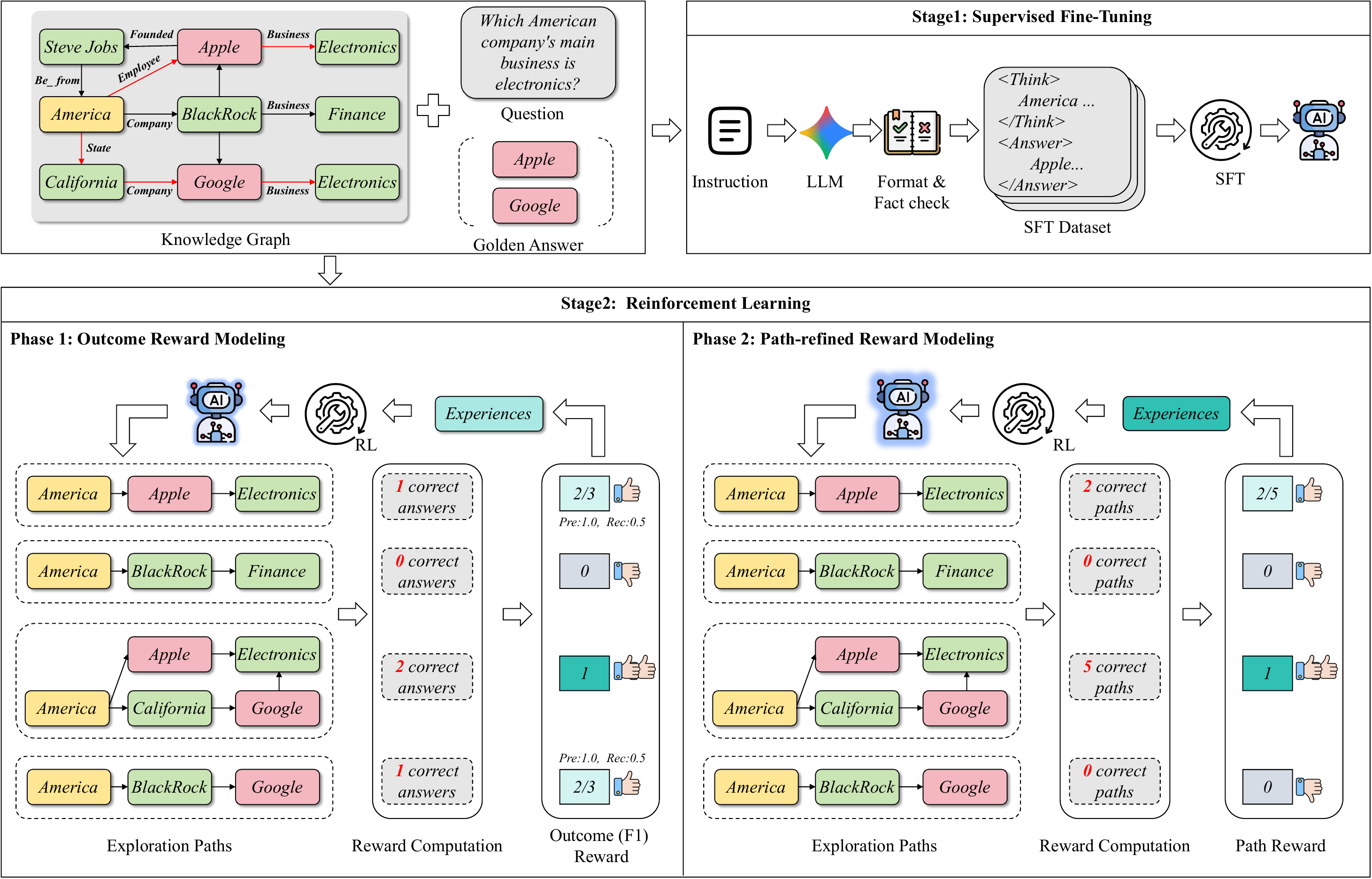}
    \caption{The overall framework of our approach. Red arrows in the knowledge graph stands for the golden reasoning paths of the given question.}
    \vspace{-7pt}
    \label{fig:framework}
\end{figure*}
In this section, we introduce our Explore-on-Graph (EOG) in detail, which consists of a Supervised Fine-Tuning (SFT) stage and a Reinforcement Learning (RL) stage.
To equip LLMs with the foundational ability for graph exploration, we introduce the SFT on knowledge graph reasoning tasks in Section \ref{subsec:longcot}.
To motivate the exploration of various reasoning paths and enhance its efficiency and meaningfulness, we introduce the RL with path-refined reward modeling in Section \ref{subsec:rl}.
The general framework of our approach is illustrated in Figure \ref{fig:framework}.

% The core objective of our method is to train a model, parameterized by $\theta$, to produce a high-quality output sequence $z$ by performing multi-step reasoning over a structed knowledge graph. Formally, given a set of inputs including a graph $\mathcal{G}$, a question $x$, and a starting entity $e_s$, the model learns to generate a sequence $z = (r, y)$ that maximizes a predefined quality objective.Our method can be formulated into three stages as detailed below.
% \subsection{Knowledge Graph Construction}
% Due to the limited input length of  language models and the substantial increase in training time as the input length grows, we opt to use a  language model to extract relevant triples from the  knowledge graph $\mathcal{G}$ that are related to the question.Given starting entities $e_s$ and a question $q$, we use a language model to extract entities related to the question from its three-hop neighbors. The triples $(h, r, t)$  formed by these entities and relations constitute the graph $\mathcal{G}_{\text{train}}$.

% \begin{equation}
% \mathcal{G}_{\text{train}} = \{ (h, r, t) \in \mathcal{G} \mid h, t \in \{e_s\} \cup \text{}_{\mathcal{M}_{\text{LM}}}(q, \mathcal{N}_3(e_s)) \}
% \end{equation}

% where $\text{Extract}_{\mathcal{M}_{\text{LM}}}$ denotes a language model-based function that extracts a set of relevant entities from the 3-hop neighborhood $\mathcal{N}_3(e_s)$ of the start entity $e_s$, conditioned on the question $q$.

\subsection{Supervised Fine-Tuning}
\label{subsec:longcot}
% Without establishing a solid foundation for the subsequent exploration phase, the reinforcement learning process would operate in an excessively vast and unstructured action space, leading to inefficient exploration, reward sparsity, and potentially misguided policy updates. 
% To mitigate these challenges, we introduce Long Chain-of-Thought Supervised Fine-tuning, which provides the model with structured, exemplary reasoning trajectories. 
% This approach not only equips the model with essential reasoning skills but also effectively constrains and guides the subsequent exploration, enabling more efficient and meaningful policy refinement.
Directly instructing the LLMs to explore KGs presents significant challenges, including an intractably large action space and extreme reward sparsity~\citep{xiong-etal-2017-deeppath,LinRX2018:MultiHopKG}, due to lack of prior knowledge.
To bridge this gap, CoT offers a pathway of structured reasoning demonstrations, by teaching LLMs the complex multi-step logic necessary for knowledge graph exploration~\citep{wu2024cotkr}.
Compared to short CoT, long CoT allows more comprehensive and in-depth reasoning by exploring a wider range of logical paths~\citep{chen2025towards}.
 % This approach explicitly trains LLMs to follow coherent reasoning paths that mirror the multi-step logic required for robust graph navigation.
Therefore,  we employ the long CoT Supervised Fine-Tuning paradigm to establish a foundation for the subsequent exploration phase, which consists of two parts: (1) constructing the long CoT reasoning dataset and (2) fine-tuning LLMs with the constructed dataset.

\textbf{Long CoT Dataset Construction.} We carefully design the prompt to generate long CoT for graph reasoning, 
which requires the reasoning process to be structured, logical and to be aligned with KGs. The detailed prompt could be found in Figure \ref{fig:appendix_sft_prompt}.
Due to its advanced deep thinking capabilities, we select Gemini 2.5 Flash~\citep{comanici2025gemini} to perform knowledge distillation to generate long-chain reasoning thoughts and the final answers.
Given a knowledge graph $\mathcal{G}$ and a question $q$ as the query input, the generation sample from Gemini 2.5 Flash can be defined as the reasoning process $z$, which comprises both the reasoning paths (within ``$<think></think>$'' tags) and the final answers (within ``$<answer></answer>$'' tags).
% Then we introduce a strict filtering process where we retain only the samples that adhere to the structured format and feature a reasoning path and the corresponding answers aligned with the knowledge graph.
Then we incorporate additional rules to filter the reasoning process that is both structurally and factually correct.
Finally, the SFT dataset $\mathcal{D}_\textrm{CoT}$ is constructed, with high-quality long CoT reasoning processes.
\textbf{Supervised Fine-Tuning.}
Following dataset construction, we perform SFT to endow the model with structured long CoT reasoning capabilities. 
% Given a knowledge graph $\mathcal{G}$ and a question $q$ as input, the model is trained to generate a complete reasoning process $z$, which comprises both the reasoning path and the final answers. 
We introduce a standard language modeling objective:
% \begin{equation}
% \mathcal{L}_{\textrm{SFT}} = - \sum_{(\mathcal{G}_i, q_i, z_i) \in \mathcal{D}_\textrm{CoT}} \sum_{j=1}^{|z_i|} \log P(z_{i,j} | z_{i,<j}, \mathcal{G}_i, q_i; \theta),
% \end{equation}
\begin{equation}
\mathcal{L}_{\textrm{SFT}} = - \frac{1}{|\mathcal{D}_\textrm{CoT}|} \sum_{i=1}^{|\mathcal{D}_\textrm{CoT}|} \sum_{j=1}^{|z_i|} \log P(z_{i,j} | z_{i,<j}, \mathcal{G}_i, q_i; \theta),
\end{equation}
where $(\mathcal{G}_i, q_i, z_i) \in \mathcal{D}_\textrm{CoT}$ is the $i$-th sample in the long CoT dataset, $z_i$ represents the complete reasoning process for the $i$-th sample, $z_{i,j}$ is the $j$-th token of that sequence, and $\theta$ denotes the LLM parameters. 
% This training stage compels the model to not only derive the correct answer but also to articulate a step-by-step, verifiable reasoning process grounded in the provided knowledge graph.
Through this phase, the LLM acquires a strong capacity for structured reasoning, establishing an essential foundation for the subsequent exploration phase. Reinforcement Learning (RL) can then effectively refine and diversify these learned pathways based on environmental feedback, rather than starting from unguided random exploration.

\subsection{Reinforcement Learning}
\label{subsec:rl}
While the SFT equips the LLM with the ability to generate structured reasoning paths, 
its outputs are usually confined to the patterns observed in the training data,
potentially leading to suboptimal reasoning. 
% To discover a broader and more effective exploration space, we introduce exploration-oriented reinforcement learning (RL) with path-refined reward modeling in this stage.
% The goal of this stage is to incentivize LLMs to explore novel entities and relations within the knowledge graph.
To facilitate the discovery of a broader and more effective exploration space, we structure our reinforcement learning approach into two distinct phases.
Inspired by advances such as Deepseek-R1~\citep{deepseek-math}, 
firstly, we introduce the Group Relative Policy Optimization (GRPO) to optimize the exploration policy on KGs based on the correctness of the reasoning path's final answers in Section~\ref{subsubsec:grpo}.
Then we integrate a path-refined reward that provides additional signals to steer exploration toward more efficient and meaningful paths in Section~\ref{subsubsec:pathreward}.
% However, its drawback is insufficient exploration capability---the model only excels at the reasoning patterns present in the long CoT data and lacks the ability to generalize. 
% Inspired by advances such as Deepseek-R1~\cite{deepseek-math}, we introduced Group Relative Policy Optimization (GRPO) to optimize the exploring policy $\pi_\theta$ toward  

% train the post-SFT model with the aim of stimulating the model's exploring capability over structured knowledge graphs. 

% Compared to SFT, RL is more like trial-and-error learning, excels at exploring new strategies, and the model adjusts its strategy (model parameters) based on correctness (reward signal). Our entire task can be formulated as $z = f(\mathcal{G}, q, e, r_g)$, where $\mathcal{G}$ is the knowledge graph, $q$ is the question, $e$ is the starting entity, and $r_g$ is the pre-defined reasoning path.
\subsubsection{Reinforcement Learning with Outcome Reward}\label{subsubsec:grpo}
% To further enhance the model's reasoning and exploration capabilities, we apply Group Relative Policy Optimization (GRPO) to refine the post-SFT model. GRPO avoids the need for a learned value function by calculating relative advantages within a group of sampled trajectories.
In the first phase, we optimize the exploration policy $\pi_\theta$ for enhancing the discovery of correct answers via a GRPO-based and outcome-supervised reinforcement learning objective $\mathcal{J}_{\textrm{GRPO}}(\theta)$ .
Given the environment $(\mathcal{G}, q) \in \mathcal{D}_\textrm{GRPO}$, our LLM after the SFT stage could generate a group of $S$ candidate exploration processes $\left \{ p_i \right \} _{i=1}^{S} \subseteq \mathcal{P}_{(\mathcal{G}, q)}$, where each $p_i$ is $i$-th process sampled from the old policy $\pi_{\theta_{\textrm{old}}}$.
Similar to a reasoning process $z$, each $p$ contains both the exploration paths and the final answers with structured tags.
The policy model could be optimized by maximizing the following reinforcement learning objective $\mathcal{J}_{\textrm{GRPO}}(\theta)$:

\begin{equation} \label{eq:grpo_detailed}
\begin{split}
    \mathcal{J}&_{\textrm{GRPO}}(\theta) = {\mathbb{E}_{[\mathcal{G},q \sim {\{\mathbb{P}(\mathcal{G},q) |(\mathcal{G},q) \in \mathcal{D}_\textrm{GRPO}\}},\left \{ p_i \right  \} _{i = 1}^{S} \sim \pi_{\theta_{\textrm{old}}}(\mathcal{P}_{(\mathcal{G}, q)}|p;\mathcal{G} )]}}  \\
 &\frac{1}{S}\sum_{i=1}^{S}  \frac{1}{|p_i|} \sum_{t = 1}^{|p_i|}\left \{  \min \left[ \psi _{\theta }(p_{i,t})\hat{A}(p_i), \text{clip} \left(  \psi _{\theta }(p_{i,t}), 1 \pm \epsilon\right) \hat{A}(p_i)  \right] - \beta \mathbb{D}_{\text{KL}}(\pi_{\theta} || \pi_{\text{ref}})\right \},
\end{split}
\end{equation}
where $ \psi _{\theta }(p_{i,t})=\frac{\pi_{\theta}(p_{i,t} | q,\mathcal{G} ,p_{i< t})}{\pi_{\theta_{\text{old}}}(p_{i,t} | q,\mathcal{G} ,p_{i< t})}$, and $ \hat{A}(p_i) =\frac{R(p_i) - \textrm{mean}\left (\left \{R(p_j) \right \}_{j=1}^{S}\right )}{\textrm{std}\left (\left \{R(p_j) \right \}_{j=1}^{S}\right )}$. 
The importance sampling ratio  $ \psi _{\theta }(p_{i,t})$ measures how much more likely the new policy $\pi_{\theta}$ is to generate $p_{i,t}$ compared to the old policy $\pi_{\theta_{\text{old}}}$.
$\hat{A}(p_i)$ is the relative normalized advantage.
$\epsilon$ and $\beta$ are hyperparameters.
$\mathbb{D}_{\text{KL}}(\pi_{\theta} || \pi_{\text{ref}})$ computes the KL-divergence between the updated policy and a reference policy.
$R(p_i)$ is the reward function that measures the quality of diverse exploration paths $\left \{ p_i \right \} _{i=1}^{S} $.

% This objective encourages high-reward and reliable exploration over KGs.
By optimizing Equation (\ref{eq:grpo_detailed}), the LLM learns to favor exploration paths that are highly rewarding and reliable relative to other sampled paths.
% , effectively exploring the solution space for complex graph-based reasoning tasks.
% This objective encourages high-reward and xx  exploration over KGs.
% This objective encourages high-reward and reliable exploration over KGs.
% \begin{itemize}
%     \item $\rho^{(i)}_{\theta}$ is the importance sampling ratio, $\frac{\pi_{\theta}(z_i|s)}{\pi_{\text{old}}(z_i|s)}$.
%     \item $A_i$ is the group-relative advantage for the $i$-th sequence.
%     \item $D_{\text{KL}}(\pi_{\theta} \,||\, \pi_{\text{ref}})$ is the KL divergence penalty.
% \end{itemize}
% The group-relative advantage $A_i$ is computed by standardizing the rewards of all $G$ sampled sequences:
% \begin{equation}
% \label{eq:advantage}
% A_i = \frac{R(z_i) - \text{mean}(\mathbf{R})}{\text{std}(\mathbf{R}) + \epsilon}, \quad \text{where } \mathbf{R} = \{R(z_1), \dots, R(z_G)\}.
% \end{equation}
% Therefore, 
% we define the reward function $R(p)$ to measure the correctness of a generated exploration sequence $p$ 
% by comparing the final answer entities $E_p$ with the ground-truth answer entities $E_t$. 
Therefore, we define the outcome reward $R(p)$ to encourage generating exploration
paths that contain the right answers.
Specifically, we utilize the entity-level F1 score to
measure the correctness of the final answers of a generated exploration sequence $p$.
We first recognize the set of predicted answer entities $A_p$ through extracting the texts within the ``$<answer></answer>$'' tags in the generated exploration sequence.
Considering the set of ground-truth answer entities $A_g$, then the outcome reward $R_{\textrm{outcome}}(p)$  is calculated as follows:
 \begin{equation}
 R_{\textrm{outcome}}(p)  = 2 \cdot \frac{Pre \cdot Rec}{Pre + Rec}, \textrm{where } Pre = \frac{|A_p \cap A_g|}{|A_p|}, \quad Rec = \frac{|A_p \cap A_g|}{|A_g|}
 \end{equation}

% we define the reward function $R(p)$ for a generated exploration sequence $p$ based on the F1 score of final answers,
% which quantifies the overlap between the predicted answer entities $A_p$ and the ground-truth answer entities $A_g$. 
% First, we define entity-level precision (P) and recall (R) as: \begin{equation} P = \frac{|A_p \cap A_g|}{|A_p|}, \quad R = \frac{|A_p \cap A_g|}{|A_g|} \end{equation} The F1 score is then calculated as the harmonic mean of precision and recall: \begin{equation} R(z) = \text{F1} = 2 \cdot \frac{P \cdot R}{P + R} \end{equation} 
% This reward function robustly handles cases with multiple answers, rewarding the model for being both accurate and comprehensive in its predictions. 
We set the reward to 0 if the predicted answer set $A_p$ is empty.
Note that the outcome reward $R_{\textrm{outcome}}(p)$ is automatically set to 0 for all samples that do not contain correctly formatted answer tags in the exploration sequence.
Therefore, this reward implicitly encourages the correct structured exploration processes during the policy optimization process. For this reason, we did not include an additional format reward in this phase.

\subsubsection{ Reinforcement Learning with Path-refined reward}\label{subsubsec:pathreward}
% In knowledge graph reasoning, reaching the correct answer is only part of the objective; the model must also follow a logical and verifiable reasoning path. To this end, we introduce a path-based reward function designed to explicitly measure the alignment between the model's generated reasoning text and the ground-truth reasoning path. The primary motivation for this design is to encourage "faithful reasoning"—a process where the model's textual explanation accurately reflects the traversal of entities and relations within the graph. This is crucial in KG scenarios as it ensures the model is not merely "hallucinating" a correct answer but is genuinely grounding its deductions in the provided structural knowledge, thereby making its reasoning process transparent and trustworthy.
The process of generating an exploration path reflects the LLM's reasoning trajectory through a KG.
Rewarding correct paths minimizes exploration of wrong directions. 
In addition, 
the exploration paths with sequential chains of connected relational triples contain rich semantic information, 
which can help the LLM generate more comprehensive and in-depth exploration processes.
In this phase, we suggest using path information to generate auxiliary rewards, thereby optimizing the exploration process and curtailing inefficient actions.

The path-based reward, $R_{\text{path}}(p)$, is formulated to quantify the proportion of ground-truth reasoning steps successfully articulated in the generated thought process.
To obtain the ground-truth reasoning path $r_g$ for datasets that do not provide them explicitly, we implement a ``Search-and-Verify'' pipeline. 
First, We identify the topic entities in the question and the ground-truth answer entities in the Knowledge Graph. 
We then perform a Breadth-First Search (BFS) to retrieve all potential paths connecting the topic entities to the answer entities within a maximum hop constraint. 
This step ensures high recall of potential reasoning chains.
Second, since BFS may yield spurious paths that are topologically connected but semantically unrelated, we employ a LLM (e.g., Gemini-2.5-Flash) to semantically verify these paths. 
The LLM is prompted to determine if a path logically corresponds to the question's intent, and only validated paths are retained as $r_g$.

Given a ground-truth reasoning path $r_g$, which consists of a set of structured triplets, we first extract this set of triplets, denoted as $T = \{ (s_i, r_i, o_i) \}_{i=1}^{|T|}$. 
For each triplet $t_i = (s_i, r_i, o_i) \in T$, we check if all three components—subject $s_i$, relation $r_i$, and object $o_i$—are present as substrings within the generated reasoning text $p_{\text{think}}$.
The reward is then calculated as the ratio of fully matched triplets to the total number of triplets in the ground-truth path: \begin{equation} R_{\text{path}}(p) = \frac{1}{|T|} \sum_{t_i \in T} \mathbb{I}(s_i \in p_{\text{think}} \land r_i \in p_{\text{think}} \land o_i \in p_{\text{think}}) \end{equation} where $\mathbb{I}(\cdot)$ is the indicator function, which returns 1 if the condition is true and 0 otherwise.
This function provides a direct and interpretable score of how faithfully the model's reasoning follows the correct path, hence contributing to the efficiency and meaningfulness of exploration process.

Furthermore, we introduce the joint reward $R_{\text{joint}}(p)$ to incentivizing the LLM to produce both high-quality exploration paths and the right final answers:
\begin{equation} \label{eq:jointreward}
R_{\text{joint}}(p) =  R_{\textrm{outcome}}(p) + \alpha  R_{\textrm{path}}(p)
\end{equation} 
where $\alpha$ is the coefficient that controls the relative influence of the outcome reward and the path reward on the exploration optimization policy.

% The path-based reward, $R_{path}(p)$, is formulated to quantify the proportion of ground-truth reasoning steps successfully articulated in the generated thought process.

% To obtain the ground-truth reasoning path $r_g$ for datasets that do not provide them explicitly (e.g., WebQSP, CWQ, GrailQA), we implement a ``Search-and-Verify'' pipeline. First, we perform a Breadth-First Search (BFS) on the KG to retrieve all potential paths connecting the topic entity (identified in the question) to the ground-truth answer entity. Second, since BFS may yield spurious paths that are topologically connected but semantically unrelated, we employ a large language model (e.g., Gemini-2.5-Flash) to semantically verify these paths. The LLM is prompted to determine if a path logically corresponds to the question's intent, and only validated paths are retained as $r_g$.

% Given a ground-truth reasoning path $r_g$, which consists of a set of structured triplets, we first extract this set of triplets, denoted as $T = \{(s_i, r_i, o_i)\}_{i=1}^{|T|}$.

\section{Experiments}
\begin{table*}[t]
\centering
\caption{Performance of EoG and previous state-of-the-art models on the five KGQA test sets. The best scores are in bold. $^\dagger$ marks the performance of the closed-source model which uses the same input as our EoG. EoG\textsubscript{\textit{SFT}} is the model which is only trained with SFT datasets but not with RL.}
\label{tab:performance_comparison}
\setlength{\tabcolsep}{2pt}
\resizebox{0.9\textwidth}{!}{
\begin{tabular}{llcccccccccc}
\toprule
\multirow{2}{*}{Method} & \multirow{2}{*}{Model} & \multicolumn{2}{c}{WebQSP} & \multicolumn{2}{c}{CWQ} & \multicolumn{2}{c}{GrailQA} & \multicolumn{2}{c}{QALD10-en} & \multicolumn{2}{c}{2WikiMultihop} \\
\cmidrule(lr){3-4}\cmidrule(lr){5-6}\cmidrule(lr){7-8}\cmidrule(lr){9-10}\cmidrule(lr){11-12}
& & Hit@1 & F1 & Hit@1 & F1 & Hit@1 & F1 & Hit@1 & F1 & Hit@1 & F1 \\
\midrule
KD-CoT~(\citeyear{wang2023knowledge}) &  Llama-2-7B & 68.6 & 52.5 & 55.7 & - & - & - & - & - & - & - \\
EWEK-QA~(\citeyear{dehghan2024ewek}) & - & 71.3 & - & 52.5 & - & 60.4 & - & - & - & - & - \\
\multirow{2}{*}{ToG~(\citeyear{sun2023thinkongraph})} & ChatGPT & 76.2 & - & 57.6 & - & 68.7 & - & 50.2 & - & - & - \\
& GPT-4 & 82.6 & - & 68.5 & - & 81.4 & - & 53.8 & - & - & - \\
RoG~(\citeyear{luo2024rog}) & Llama-2-7B & 85.7 & 70.8 & 62.6 & 56.2 & - & - & - & - & - & - \\
ODA~(\citeyear{sun-etal-2024-oda}) & GPT-4 & - & - & - & - & - & - & 66.7 & - & - & - \\
EffiQA~(\citeyear{dong2025effiqa}) & GPT-4 & 82.9 & - & 69.5 & - & 78.4 & - & 51.4 & - & - & - \\
GNN-RAG~(\citeyear{mavromatis2025gnn}) & Llama-2-7B & 85.7 & 71.3 & 66.8 & 59.4 & - & - & - & - & - & - \\
KG-Agent~(\citeyear{jiang-etal-2025-kg}) & Llama-2-7B & 83.3 & 81.0 & 72.2 & 69.8 & - & 86.1 & - & - & - & - \\
\multirow{2}{*}{DoG~(\citeyear{li-etal-2025-decoding-graphs})} & Qwen2.5-7B & 92.7 & - & 74.1 & - & - & - & - & - & 84.2 & - \\
& Llama-3.1-8B & 91.4 & - & 76.2 & - & - & - & - & - & 84.1 & - \\
GCR~(\citeyear{luo2024graph}) 
& Llama-3.1-8B  & 92.2 & 79.1 & 75.8 & 61.7 & - & - & - & - & - & - \\
\midrule
$^\dagger$Gemini-2.5 Flash\label{row:Gemini-2.5 Flash} & - & 91.8 & 78.2 & 65.5 & 59.3 & 90.3 & 83.8 & 56.7 & 46.2 & 83.9 & 83.1  \\
$^\dagger$Gemini-2.5 Pro\label{row:Gemini-2.5 Pro} & - & 92.1 & 79.8 & 71.9 & 65.3 & 91.6 & 84.5 & 58.6 & 48.3 & 85.1 & 82.6 \\
$^\dagger$GPT-5\label{row:GPT5} & - & 86.1 & 77.5 & 74.1 & 67.6 & 90.5 & 85.4 & 59.2 & 50.4 & 84.2 & 83.4  \\
\midrule
\multirow{2}{*}{R\textsubscript{\textit{quality}}} & Qwen2.5-7B  & 83.9 & 72.6 & 68.3 & 60.3 & 89.2 & 87.6 & 55.6 & 44.1 & 82.5 & 81.9 \\
& Llama-3.1-8B & 86.3 & 74.5 & 70.5 & 62.1 & 91.4 & 88.2 & 57.1 & 48.7 & 83.1 & 82.7 \\
% EoG\textsubscript{\textit{SFT}}   & Llama-3.1-8B & 84.3 & 74.3 & 72.1 & 63.2 & 91.5 & 86.6 & 91.5 & 86.6 & 84.9 & 82.4 \\
% \textbf{EoG} & Llama-3.1-8B & \textbf{92.8} & \textbf{80.7} & \textbf{84.3} & \textbf{73.9} & \textbf{96.1} & \textbf{91.6} & 96.1 & 91.6 & \textbf{85.3} & \textbf{84.3} \\
\multirow{2}{*}{EoG} & Qwen2.5-7B  & 90.7 & 78.1 & 82.7 & 73.8 & 91.7 & 88.5 & 67.3 & 57.8 & 83.9 & 82.9 \\
& Llama-3.1-8B & \textbf{92.8} & \textbf{81.3} & \textbf{86.6} & \textbf{77.9} & \textbf{92.1} & \textbf{90.6} & \textbf{70.6} & \textbf{61.9} & \textbf{85.3} & \textbf{84.3} \\
\bottomrule
\end{tabular}
}
\end{table*}

\subsection{Experiments Setup}
\textbf{Datasets.} Following previous research, we evaluate our method on five widely-used benchmark datasets: CWQ~\citep{talmor-berant-2018-web}, WebQSP~\citep{yih-etal-2016-value}, GrailQA~\citep{Gu_2021}, QALD10-en~\citep{usbeck2024qald} and 2WikiMultihop~\citep{ho2020constructingmultihopqadataset}. The first three benchmarks are constructed based on Freebase, whereas QALD10-en and 2WikiMultihop are built upon Wikidata. The details of the datasets are
described in Appendix \ref{app:datasets}.

\textbf{Baseline.} We compare EoG with 10 previous baseline methods~\citep{luo2024rog,sun-etal-2019-pullnet}, such as GCR and DoG. The detailed descriptions of the baseline methods are listed in Appendix 
\ref{app:baselines}. 
In particular, we report the performance of closed-source LLMs with strong deep-thinking ability, Gemini 2.5 series~\citep{comanici2025gemini} and GPT-5\footnote{https://cdn.openai.com/gpt-5-system-card.pdf},
by querying the APIs with the same instructions and inputs as our EoG, including the questions and the corresponding KGs.
The implementation details are provided in Appendix \ref{app:details}.

\textbf{Evaluation Metrics.} In the evaluation of baselines, we adopt the Hit@1 and F1 metrics following previous work~\citep{mavromatis2025gnn}. The Hit@1 metric focuses on whether any correct answer is present in the prediction, while the F1 metric evaluates the coverage of all answers by comprehensively considering the precision and recall of the prediction.

\textbf{Implementations.}
To validate the general applicability of our approach, we apply EoG to two open-source LLMs, Qwen2.5-7B-Instruct ~\citep{yang2024qwen2technicalreport} and Llama-3.1-8B-Instruct ~\citep{dubey2024llama}. For fine-tuning, we use Gemini-2.5-Flash to generate long chain-of-thought datasets. For reinforcement learning, we implement the GRPO method using verl ~\citep{Sheng_2025}. More details on experimental settings can be found in Appendix \ref{experiment}.
% To ascertain the general applicability of our approach, we apply EOG to two opensource LLMs,Qween-2.5-7b-Instruct~\citep{yang2024qwen2technicalreport} and Llama-3.1-8b-Instruct~\citep{dubey2024llama}. For finetuning,we use Gemini-2.5-Flash to achieve Long cot datasets. For reinforcement learining ,we implement the grpo method using verl. More details on experimental settings can be found in Appendix B.
% During inference, we leveraged the vLLM~\cite{kwon2023efficientmemorymanagementlarge} backend with a tensor parallelism degree of 2, temperature set to 1, top-$p$ to 1, and generated 8 responses per prompt.

\subsection{Main Results}
From Table \ref{tab:performance_comparison} we have several observations.
First, our EoG outperforms previous KG-enhanced reasoning methods with similar amounts of LLM parameters. 
Impressively, it even exceeds the performance of existing closed-source deep-thinking LLMs such as Gemini 2.5 Pro and GPT-5.
It demonstrates the remarkable  strength of EoG's autonomous exploration capability.
Second, the performance of EoG\textsubscript{\textit{SFT}} significantly drops on all datasets compared with EoG,
which indicates that EoG's strong exploration capability is attributed to the reinforcement learning stage, rather than knowledge distillation from LLMs, further demonstrating the effectiveness of our proposed RL with path-refined reward modeling.
Finally, EoG built upon two base models both demonstrate strong performance, proving the general applicability of our method to different LLM architectures.

% achieves state-of-the-art results across multiple datasets, which indicates the effectiveness of our framework on knowledge graph reasoning tasks. 
% Moreover, while the EoG (w/o RL) variant does not surpass the previous baseline, its performance after reinforcement learning training reaches state-of-the-art levels. 
% It indicates  that SFT(Supervised Fine-tuning) is a necessary preceding step by stimulate exploration capabilities.
% Finally, EOG outperforms GPT-5 and other leading models, demonstrating that our approach enables a 7B parameter model to achieve performance competitive with much larger, commercial closed-source large models, further highlighting the effectiveness and practicality of our framework.
\subsection{Ablation Study}
\textbf{Ablation Study.} We conduct an ablation study on the CWQ and WebQSP test sets to validate the contribution of each component in our framework. 
The results are shown in Table \ref{tab:ablation_combined}.
% demonstrate that all the components significantly contribute our model.
Note that when removing the outcome reward, we use the Hit@1 score instead to reward the exploration process.
First, we observe that the path reward is critical to performance, demonstrating that our path-refined reward effectively enhances EoG's exploration by optimizing the semantic meaningfulness of its reasoning paths.
Furthermore, the RL phase with outcome reward constitutes
a significant contribution to EoG's performance.
This shows that a granular evaluation of the answers plays an important role in the process of exploring policy optimization, demonstrating the necessity of using the F1 score.
% confirming F1's superiority as a more effective training signal.
Finally, EoG without SFT, which is only trained with RL, 
shows suboptimal performance, regardless of whether the In-Context Learning (ICL) is applied, 
proving the necessity of adopting SFT 
for the cold start in our framework.

\begin{table}[h]
    \centering
    \caption{Ablation studies of EoG on CWQ and WebQSP datasets. }
    \label{tab:ablation_combined}
    \resizebox{0.6\textwidth}{!}{
    \begin{tabular}{l|cc|cc}
        % \multicolumn{5}{c}{(a) Reward} \\
        \hline
        & \multicolumn{2}{c|}{CWQ} & \multicolumn{2}{c}{WebQSP} \\
        Model & Hit@1 & F1 & Hit@1 & F1 \\
        \hline
        EoG & \textbf{82.6} &\textbf{73.9} & \textbf{92.8} & \textbf{81.3} \\
        \hline
        \ \ \ \ w/o path reward & 81.5 & 70.8 & 90.2 & 77.3 \\
        \ \ \ \ w/o outcome reward& 62.7 & 51.4 & 65.5 & 56.2 \\
        % Full Model & \textbf{84.3} &\textbf{73.9} & \textbf{92.8} & \textbf{80.7} \\
        \ \ \ \ w/o  SFT & 70.3 & 63.1 & 75.9 & 65.8 \\
        \ \ \ \ w/o  SFT, w/ ICL & 70.7 & 63.8 & 77.2 & 66.5 \\
        \hline
    \end{tabular}
    % \hspace{2em}
    % \begin{tabular}{l|cc|cc}
    %     % \multicolumn{5}{c}{(b) SFT} \\
    %     \hline
    %     & \multicolumn{2}{c|}{CWQ} & \multicolumn{2}{c}{WebQSP} \\
    %     Configuration & Hit@1 & F1 & Hit@1 & F1 \\
    %     \hline
    %     Full Model & \textbf{84.3} &\textbf{73.9} & \textbf{92.8} & \textbf{80.7} \\
    %     Pure RL & 70.3 & 63.1 & 75.9 & 65.8 \\
    %     Pure RL+ICL & 70.7 & 63.8 & 77.2 & 66.5 \\
    %     \hline
    % \end{tabular}
    }

\end{table}

\begin{figure}[t]
    \centering
    % 左侧子图
    \begin{minipage}[t]{0.50\textwidth}
        \centering
        \includegraphics[width=0.95\linewidth]{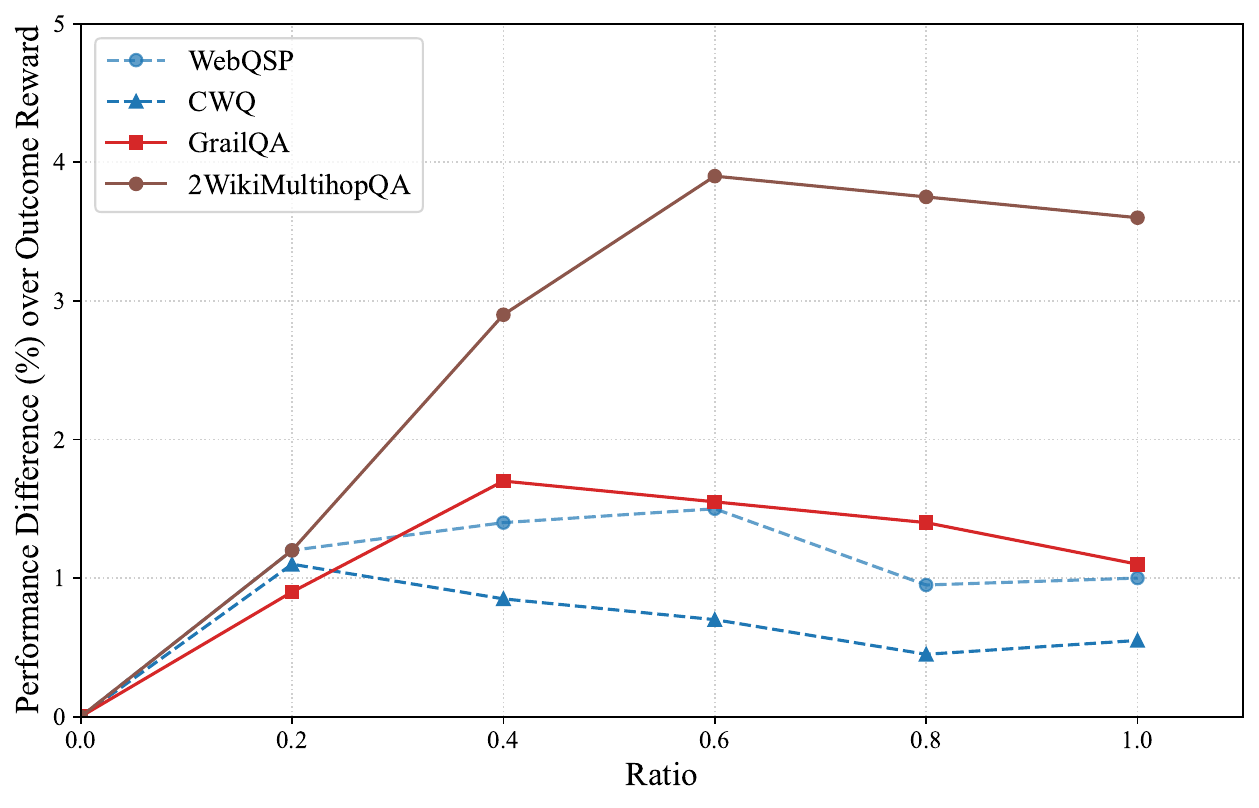}
        \captionof{figure}{Performance of EoG with different ratios of the path reward to the outcome reward. In
the figure, the horizontal axis represents the differ-
ent ratios, while the vertical axis indicates the per-
formance difference  compared to using only
the outcome reward. }
        \label{fig:ablation_study}
    \end{minipage}%
    \hfill
    % 右侧子图
    \begin{minipage}[t]{0.48\textwidth}
        \centering
        \includegraphics[width=0.8\linewidth]{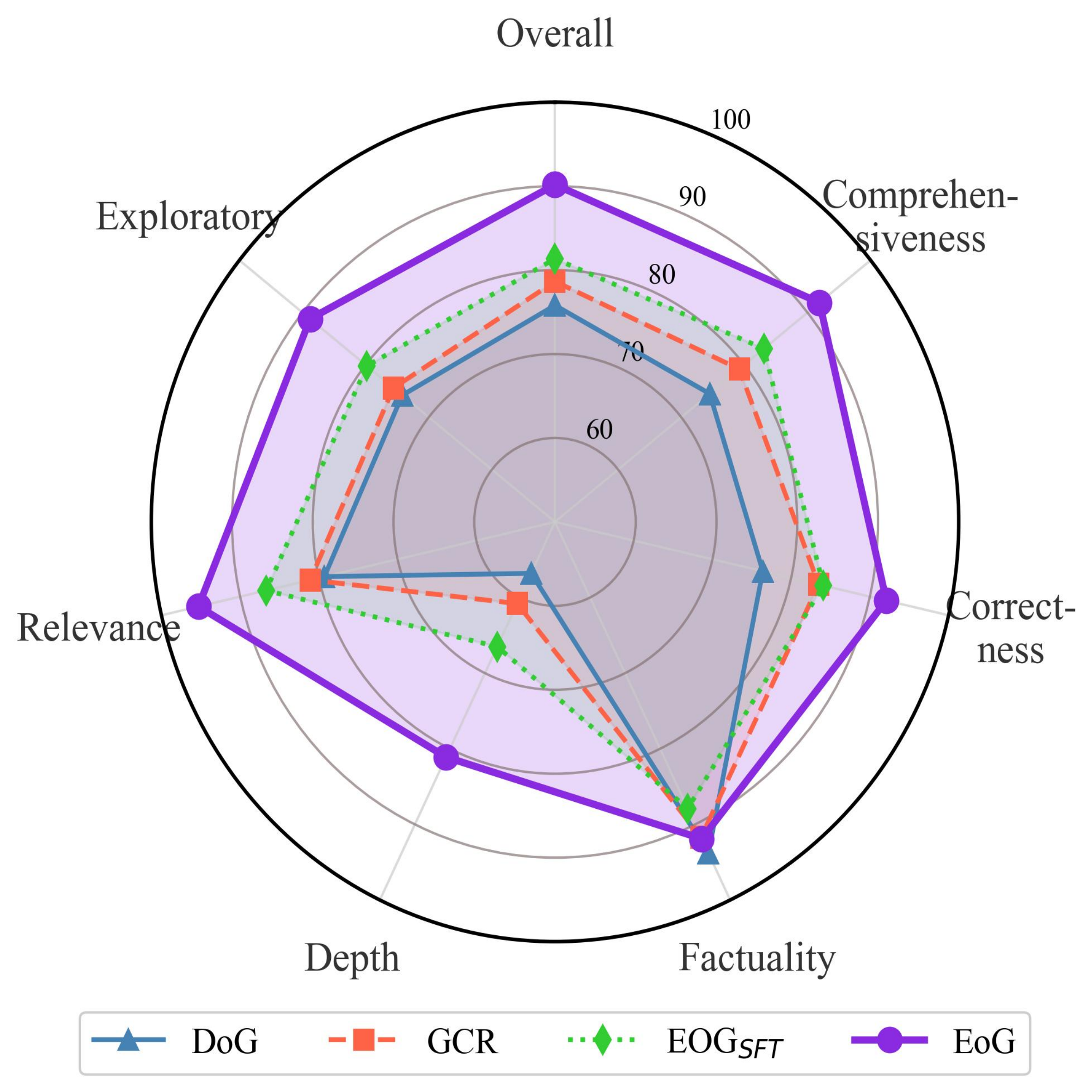}
        \captionof{figure}{A multi-dimensional performance comparison on different reasoning paths.}
        \label{fig:example-wrap}
    \end{minipage}
    \vspace{-7pt}
\end{figure}

\textbf{Balance between Outcome and Path Reward.}
To investigate the optimal balance between the outcome and path reward, we evaluated various ratios $\alpha$ in Equation (\ref{eq:jointreward}), which is illustrated in Figure \ref{fig:ablation_study}.
Note that the horizontal axis represents the different ratios, while the vertical axis indicates the performance difference (\%) compared to using only the outcome reward.
We observe that when the ratio is too small, the performance drops significantly.
We hypothesize that reducing path signals may lead EoG to generate error or meaningless paths and hurt the performance.
When the ratio is excessively increased, EoG may pay less attention to the correctness of the answers given the exploration paths, which leads to performance degradation.

\begin{table}[ht]
\centering
\caption{Evaluation of the impact of the path reward.}
\label{tab:efficiency_path_reward}
\resizebox{0.9\textwidth}{!}{
\begin{tabular}{llcccc}
\toprule
Dataset & Setting & Average Output Length($\downarrow$) & Comprehensiveness($\uparrow$) & Relevance($\uparrow$) & Exploration($\uparrow$)  \\
\midrule
\multirow{2}{*}{CWQ}
  & EoG   & \textbf{1528} & \textbf{91.9}    & \textbf{95.3} & \textbf{88.8}  \\
  & EoG w/o path reward & 2067 & 89.8 & 92.6 & 85.1  \\
\midrule
\multirow{2}{*}{WebQSP}
  & EoG   & \textbf{851} & \textbf{92.7}    & \textbf{97.1} & \textbf{78.9}  \\
  & EoG w/o path reward  & 926 & 91.9 & 95.3  & 74.3  \\
\bottomrule
\end{tabular}}
\end{table}

% \begin{figure}[h]
%     \centering
%     \includegraphics[width=0.47\linewidth]{figures/size_curve_3.pdf}
%     \caption{Placeholder for Figure 2: Ablation Study Results}
%     \label{fig:ablation_study}
% \end{figure}
% \begin{figure}[h]
%     \centering
%     \includegraphics[width=0.47\linewidth]{figures/dim.pdf}
%     \caption{Reasoning}
%     \label{fig:example-wrap}
% \end{figure}
\textbf{Impact of Path Reward.}
% We evaluated the impact of path reward based on the average output length and the 
% on the efficiency and meaningfulness of the EoG outputs.
% We evaluate efficiency based on the length of the model's output, and assess meaningfulness using A, B, and C
We conduct experiments to measure the impact of path reward on EoG's output in terms of its average length, comprehensiveness, relevance, and exploration, with the latter three metrics being assessed by querying GPT-4o-mini~\citep{hurst2024gpt} using custom-designed prompts.
The relevant prompts are shown in Figure \ref{fig:appendix_score_prompt}.
% The results are shown in Table \ref{tab:efficiency_path_reward}.
% This suggests that the path reward refined from the knowledge graph can impose constraints during EoG's exploration process, thereby enhancing reasoning efficiency.
% Additionally, the model's testing time also decreased due to the reduction in output tokens. 
From the Table \ref{tab:efficiency_path_reward}, it is demonstrated that our path reward method can improve the efficiency and meaningfulness of our EoG's output.
\subsection{Analysis of Reasoning Quality}
In this section, we evaluate the quality of the reasoning process of different methods across six dimensions.
We select GPT-4o-mini as the judge and the relevant prompts are shown in Figure \ref{fig:appendix_score_prompt}.
The results show that EoG performs outstandingly compared to other methods, indicating that the model can effectively integrate knowledge graph information to generate comprehensive and accurate answers.
Notably, our model has achieved significant improvements in reasoning depth, exploration, and comprehensiveness, demonstrating the substantial enhancement of the two-stage reinforcement learning method we employed on the exploration policy.
 % Notably, when we introduce reinforcement learning into the EoG model, the performance shows significant improvement in the depth and exploration dimensions. This improvement is attributed to the fact that reinforcement learning encourages the model to explore more relevant and in-depth knowledge during the reasoning process, rather than merely satisfying the finding of a correct answer. In contrast, other imitation-based methods, such as DoG and GCR, perform poorly in most dimensions, particularly in depth and exploration, which directly reflects the limitations of imitation-based methods in mining deep logic. These results collectively confirm that our EoG significantly enhances the quality and richness of reasoning results through effective exploration and in-depth logical analysis in knowledge graph reasoning tasks.
\subsection{Performance on Complex Reasoning Scenarios}
Following previous work~\citep{luo2024graph},
we divided the test set of the CWQ dataset according to different logical patterns and reasoning depths.
As shown in Table \ref{tab:causal_hop_combined}, we observe that EoG outperforms both GCR and DoG in almost all subsets,
especially in the more challenging reasoning scenarios, such as reasoning over superlative patterns or $\geq3$ hops.
We claim that this is because EoG actively explores a larger reasoning space on the graphs, which gains more logical and semantic information to help understand various logical patterns and deeper reasoning.
In general, the results on different causal patterns and hops show the effectiveness of our method in complicated reasoning scenarios.                         
\begin{table}[h]
\centering
\caption{Performance (F1 scores) comparison on different logical patterns and reasoning depths.}
\label{tab:causal_hop_combined}
\resizebox{0.97\textwidth}{!}{
\begin{tabular}{lllcccccccccc}
\toprule
\multicolumn{3}{c}{ Method} & Conjuction & Comparative & Superlative & Composition & 1-hop & 2-hop & 3-hop & $\geq$4-hop & Overall \\
\midrule
\multirow{6}{*}
  & \multirow{2}{*}{GCR}     & Hit@1 & 73.1 & 66.3 & 64.4 & 66.8  & 79.0 & 71.9 & 62.1 & 47.7 & 68.2 \\
  &                                   & F1 & 63.7 & 57.7 & 52.6 & 59.7  & 66.3 & 63.0 & 56.6 & 45.8 & 60.3 \\
  \hline
  & \multirow{2}{*}{DoG}     & Hit@1 & 72.3 & 77.8 & 68.7 & 75.9  & 75.1 & 74.1 & 69.9 & 63.3 & 73.7 \\
  &                                  & F1 & 53.3 & 68.3 & 45.9 & 49.2  & 50.3 & 53.7 & 64.5 & 46.7 & 53.2 \\
    \hline
  & \multirow{2}{*}{EoG}  & Hit@1 & \textbf{77.2} & \textbf{85.2} & \textbf{73.4} & \textbf{82.8}  & \textbf{83.5} & \textbf{79.1} & \textbf{83.2} & \textbf{76.8} & \textbf{82.6} \\
  &                                   & F1 & \textbf{70.2} & \textbf{77.2} & \textbf{64.7} & \textbf{76.8}  & \textbf{76.2} & \textbf{72.0} & \textbf{78.1} & \textbf{69.6} & \textbf{73.9} \\
  
\bottomrule
\end{tabular}
}
\end{table}

% \begin{table}[t]
%     \centering
%     \caption{Ablation studies on CWQ and WebQSP. Left: reward components. Right: SFT necessity.}
%     \label{tab:ablation_combined}
%     \resizebox{0.5\textwidth}{!}{
%     \begin{tabular}{l|cc|cc}
%         \multicolumn{5}{c}{(a) Reward} \\
%         \hline
%         & \multicolumn{2}{c|}{CWQ} & \multicolumn{2}{c}{WebQSP} \\
%         Model & Hit@1 & F1 & Hit@1 & F1 \\
%         \hline
%         EoG & \textbf{84.3} &\textbf{73.9} & \textbf{92.8} & \textbf{80.1} \\
%         w/o Path reward & 83.5 & 72.8 & 91.2 & 79.3 \\
%         w/o RL & 69.8 & 60.6 & 84.3 & 74.3 \\
%         w/o F1 & 32.4 & 21.7 & 35.2 & 26.1 \\
%         \hline
%     \end{tabular}
%         }
% \end{table}

\subsection{Analysis of Exploration Behavior}
\label{sec:exploration_analysis}

To examine how reward design influences exploration behavior, we evaluate models using exploration efficiency and coverage of the reasoning space on the CWQ test set.
Exploration efficiency measures the average number 
of explored triples required to identify one correct reasoning triple, 
while reasoning coverage quantifies the proportion of ground-truth reasoning triples 
successfully discovered. Detailed metric definitions are provided in Appendix~\ref{appendix:exploration_metrics}.

\begin{table}[ht]
\centering
\caption{Comparison of exploration efficiency and reasoning coverage on the CWQ test set.}
\label{tab:exploration_results}
\begin{tabular}{lcc}
\toprule
\textbf{Model} & \textbf{Exploration Efficiency ($\downarrow$)} & \textbf{Coverage ($\uparrow$)} \\ \midrule
$\text{EoG}_{\text{SFT}}$      & \textbf{2.877} & 0.615 \\
$\text{EoG}_{\text{Outcome}}$  & 3.028          & 0.689 \\
EoG           & 2.887          & \textbf{0.723} \\ \bottomrule
\end{tabular}
\end{table}

As shown in Table~\ref{tab:exploration_results}, EoG achieves the 
highest reasoning coverage while maintaining efficiency comparable to the 
SFT baseline. Although outcome-based RL improves coverage, it introduces 
noisy exploration. In contrast, our path-refined reward encourages broader 
yet precise reasoning, leading to better coverage without increasing 
exploration overhead.

\subsection{Performance on Out of Distribution Datasets}

\begin{figure}[H]
    \centering
    \begin{subfigure}[t]{0.23\textwidth}
        \includegraphics[width=\textwidth]{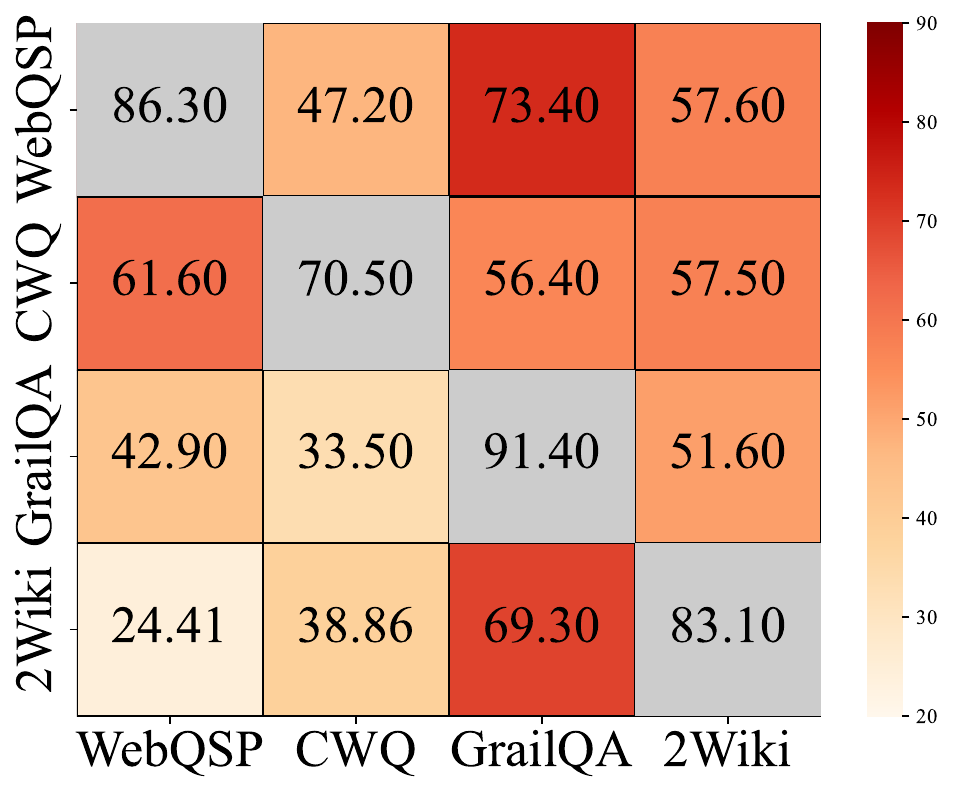}
        \caption{EoG\textsubscript{\textit{SFT}}(Hit@1)}
        \label{fig:sub1}
    \end{subfigure}%
    \begin{subfigure}[t]{0.23\textwidth}
        \includegraphics[width=\textwidth]{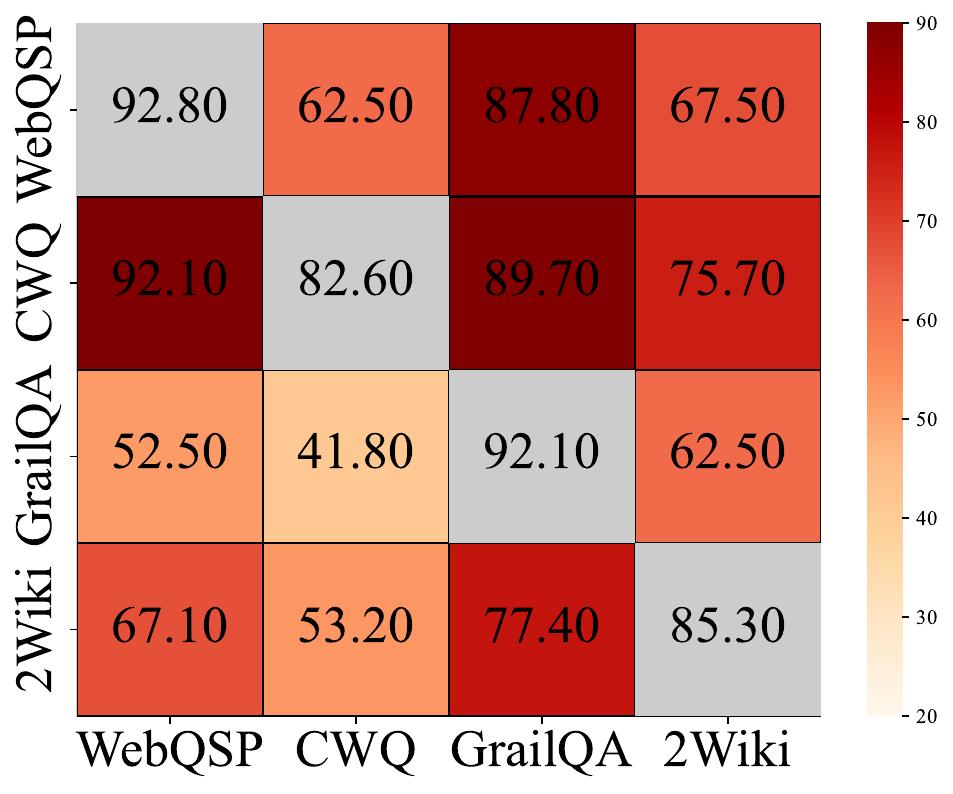}
        \caption{EoG(Hit@1)}
        \label{fig:sub2}
    \end{subfigure}%
    \begin{subfigure}[t]{0.23\textwidth}
        \includegraphics[width=\textwidth]{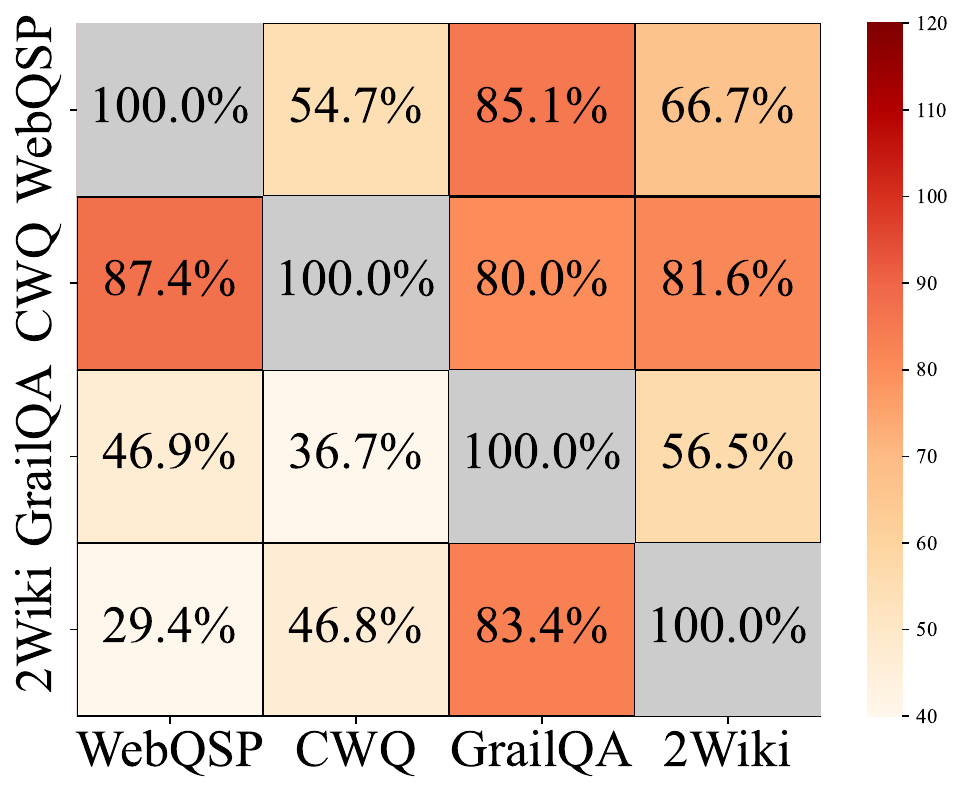}
        \caption{EoG\textsubscript{\textit{SFT}}(ratio)}
        \label{fig:sub3}
    \end{subfigure}%
    \begin{subfigure}[t]{0.23\textwidth}
        \includegraphics[width=\textwidth]{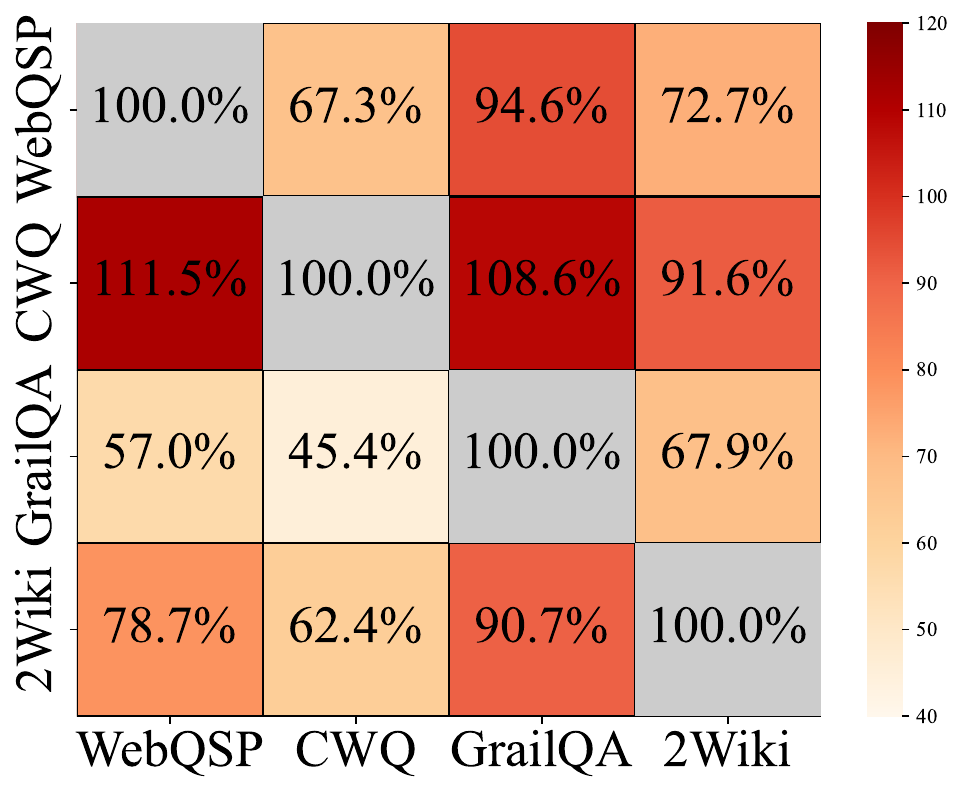}
        \caption{EoG(ratio)}
        \label{fig:sub4}
    \end{subfigure}
    \caption{Hit@1 comparison and performance ratios across four datasets under out-of-distribution (O.O.D.) settings. The O.O.D.-to-I.I.D. ratio is defined as a model's performance score on Out-of-Distribution (O.O.D.) data divided by its performance score on Independent and Identically Distributed (I.I.D.) data. Subfigures (c) and (d) show the O.O.D.-to-I.I.D. ratios of the models on the four datasets. 2Wiki refers to the 2WikiMultihop dataset. The Y-axis shows the dataset the model was trained on and the X-axis shows the dataset the model was evaluated on.}
    \label{fig:four_pdfs}
\end{figure}

Figures \ref{fig:four_pdfs}(a) and \ref{fig:four_pdfs}(b) show that EoG outperforms EoG\textsubscript{\textit{SFT}} on four datasets in O.O.D. settings.
Leveraging the autonomous exploration on graphs, EoG reliably maintains model performance stability across diverse datasets and data sources.
Figures \ref{fig:four_pdfs}(c) and \ref{fig:four_pdfs}(d) show that EoG achieves higher O.O.D.-
to-I.I.D. ratios than EoG\textsubscript{\textit{SFT}}, reflecting its
strong robustness and cross-domain generalizability via reinforcement learning with path-refined reward over KGs. 
\begin{figure}[t] % 或 [htbp]
    \centering
    % 若需要裁剪，调整 trim=left bottom right top 的 pt/mm；示例裁掉左右各5pt，上下各10pt
    \includegraphics[width=0.9\linewidth]{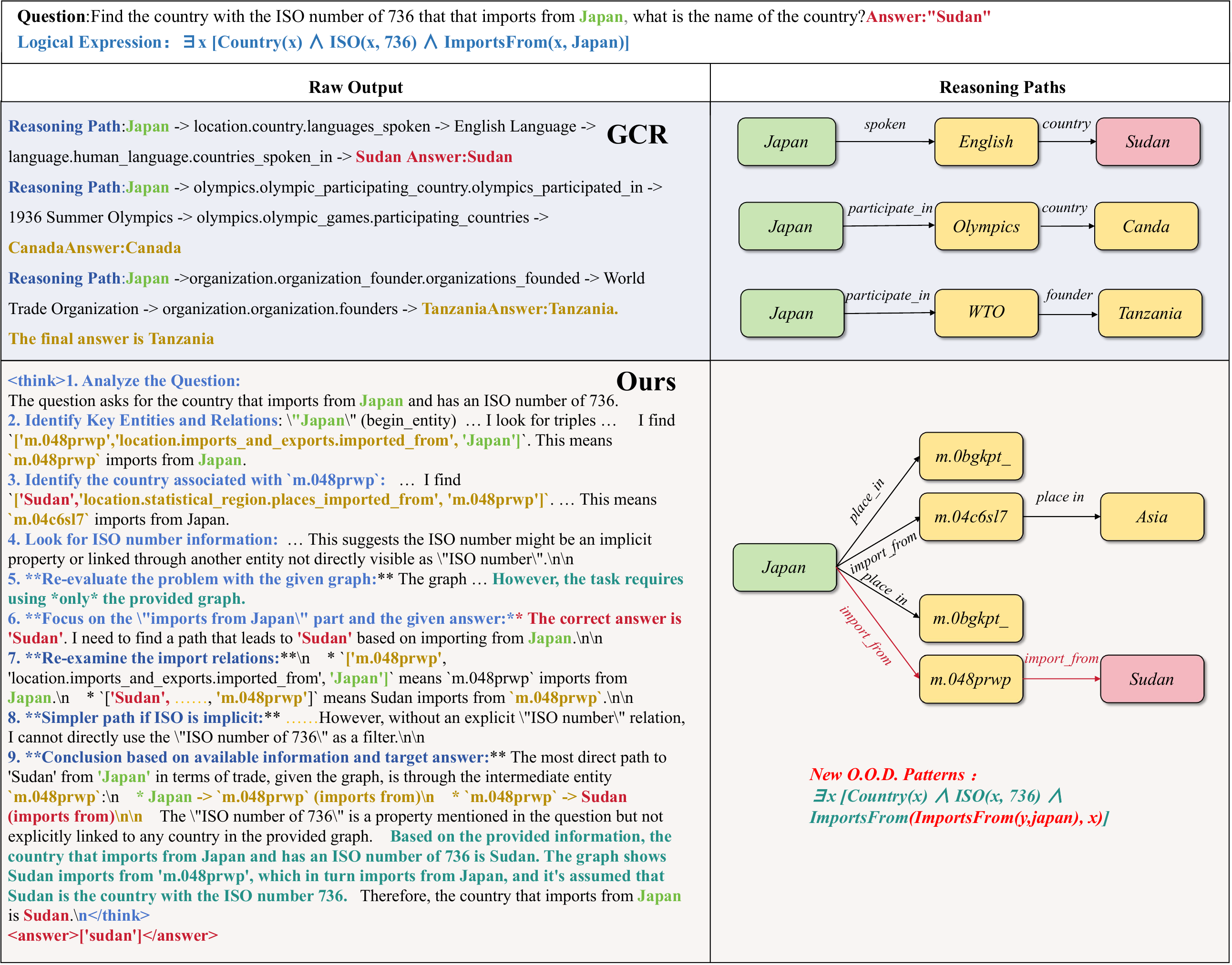}
    \caption{An example of question and the corresponding answer from EoG and GCR. Different types of entities are distinguished with different colors.}
    \label{fig:external-pdf}
\end{figure}

\subsection{Case Study}
Figure \ref{fig:external-pdf} illustrates the comparison of EoG and GCR on the knowledge graph reasoning task instance. Although the question presupposes the relational pattern ``$\exists x,   \text{ImportsFrom}(x, \text{Japan}) $'', active exploration of our model uncovered an alternative and unseen pattern: ``$\exists x ,  \text{ImportsFrom}(\text{ImportsFrom}(y, \text{Japan}), x)$'' through exploring the unknown entity ``m.048prwp''.
Due to the lack of prior knowledge, GCR fails to recognize this pattern.
This shows EoG's capacity to adapt to unexpected causal patterns through exploration.
Furthermore, in case some facts are missing in the graph, such as the label of ``m.048prwp'', the relation ``ISO'', our model can still leverage known semantic information for inference and prediction, demonstrating the strong capability of path-based policy optimization.
The above observations demonstrate the effectiveness of our method in autonomous exploration of knowledge graphs. 

% —a capability absent in GCR, which failed to identify this structure. Furthermore, even though the relationship "Japan imports from Sudan" is absent from the graph, our model hypothesized its possibility, underscoring its robust reasoning ability in the face of incomplete knowledge graphs. 
% In the second instance, when faced with the two sets requiring cross-validation—"states through which the Colorado River flows" and "states contained in the Mountain Time Zone"—the model automatically performed multi-step queries and logical intersection operations, precisely outputting the states that satisfy both conditions. This indicates that through exploration, the model discovered a pattern \(\exists x\, \exists y\, [\, \text{State}(x) \land \text{partiallyContainedBy}(\text{ColoradoRiver}, x) \land \text{timeZones}(x, y) \land y = \text{MountainTimeZone}\,]\) that goes beyond the original causal pattern \(\exists x\, [\, \text{State}(x) \land \text{RunsThrough}(\text{ColoradoRiver}, x) \land \text{InTimeZone}(x, \text{MountainTimeZone})\,]\).
% These examples demonstrate that our model exhibits strong multi-hop reasoning and complex constraint handling capabilities, whereas traditional methods are typically limited to direct relationship matching and cannot handle such implicit, combinatorial logical relationships. The above results validate the effectiveness of our training approach in inducing the model to perform deep structured reasoning.
\section{Conclusion}
In this paper, we propose a novel framework called
Explore-on-Graph (EoG) to address the critical challenge of limited generalizability in existing KG-enhanced LLM reasoning methods, which often fail on out-of-distribution reasoning patterns.
EoG employs reinforcement learning, guided by both the correctness of the final answer and a path-refined reward for efficiency, to incentivize the model's exploration of novel and meaningful reasoning paths.
Furthermore, extensive experiments demonstrate that EoG achieves new state-of-the-art results, significantly outperforming existing open-source models and even surpassing the performance of powerful closed-source LLMs. Our analysis also provides key insights into RL training strategies for KGQA. 
Finally, future work could investigate how to leverage knowledge graphs to explore more sophisticated reward strategies, thereby enabling more efficient and meaningful reasoning.

\section*{Acknowledgement}

This work was supported by Zhongguancun Laboratory, the Henan Provincial Major Industrial Key Technology Research Open Bidding for Leading Talents Project (251000210400) and Ant Group Research Fund.

\section*{Ethics Statement}

The data used in this paper is sourced exclusively from publicly available, open-source KGQA datasets (see Appendix~\ref{app:datasets}). As these datasets do not contain any personally identifiable information or other sensitive data, our experiments raise no apparent ethical concerns related to data privacy or human subjects.

\bibliography{iclr2026_conference}
\bibliographystyle{iclr2026_conference}

\appendix
\section{preliminary}
\label{app:preliminary}
\textbf{Knowledge Graphs (KGs).} A knowledge graph consists of a large number of relational triples: $\mathcal{G} = \{ (s, r, o) \mid s, o \in \mathcal{E}, r \in \mathcal{R} \}$, 
where $\mathcal{E}$ and $\mathcal{R}$ denote the set of entities and relations, respectively.
A relational triple $(s,r,o)$ represents a factual statement where the subject $s$ is connected to the object $o$ through the relation $r$.

\textbf{Reasoning Paths.} A reasoning path in the KG is a sequential chain of connected relational triples that provides a coherent and interpretable pathway from a starting entity to a target entity, denoted as:
$P = e_0 \xrightarrow{r_1} e_1 \cdots \xrightarrow{r_L} e_L$
where each step $(e_{i-1}, r_i, e_i)$ is a valid triple in the graph $\mathcal{G}$.

% \textbf{Problem Formulation.} Knowledge Graph Question Answering (KGQA) focuses on reasoning over KGs to find answers to natural language questions. 
\textbf{Problem Formulation.} In this work, our goal is to formulate a function $f$ that predicts the correct answers $a$ from the graph: $a = f(q,\mathcal{G})$, given a question $q$ and an available KG denoted by $\mathcal{G}$.
We assume that the entities mentioned in the question and the answers are all linked to their corresponding entities in $\mathcal{G}$, consistent with previous work~\citep{sun-etal-2019-pullnet,luo2024rog}.
% \textbf{problem } 

\section{Datasets}
\label{app:datasets}

\textbf{WebQSP.}
WebQSP is a foundational KGQA benchmark designed to evaluate a model's ability to answer simple, fact-based questions that typically require retrieving a single fact from the knowledge graph. The dataset provides full SPARQL queries for its questions, which are executed against \textbf{Freebase}.

\textbf{CWQ.}
CWQ extends the complexity beyond WebQSP by introducing questions that require compositional reasoning. These questions often involve multiple constraints, conjunctions, and superlatives, necessitating multi-hop or multi-relation inference paths on the \textbf{Freebase} knowledge graph. The dataset is annotated with complex SPARQL queries that reflect these reasoning structures.

\textbf{2WikiMultihopQA.}
The 2WikiMultihopQA dataset is specifically designed to assess multi-hop reasoning capabilities by integrating both structured knowledge from \textbf{Wikidata} and unstructured text from Wikipedia articles. A key feature of this dataset is the inclusion of structured 'evidences' —a sequence of triples that explicitly outlines the reasoning path from the question's entities to the answer. This makes it particularly suitable for evaluating the faithfulness of a model's generated reasoning chains.

\textbf{GrailQA.}
GrailQA is a large-scale, high-quality dataset built on \textbf{Freebase} that is designed to evaluate the generalization capabilities of KGQA models across three distinct levels: i.i.d., compositional, and zero-shot. This structure allows for a fine-grained analysis of a model's ability to handle previously seen patterns (i.i.d.), new combinations of seen patterns (compositional), and entirely new domains and relations (zero-shot).

% \begin{table}[h!]
% \centering
% \caption{Statistics of fine-tuning datasets for EoG}
% \label{tab:2wiki_stats}
% \begin{tabular}{@{}c|c|c@{}}
% \toprule
% Total & \#WebQSP & \#CWQ  \\ \midrule
% 209,909 & 28,307 & 181,602 \\ \bottomrule
% \end{tabular}
% \end{table}

\textbf{QALD-10-en.}
QALD-10 marks a significant shift by using \textbf{Wikidata} as its primary knowledge source. QALD-10-en is its English subset, which contains high-quality, complex questions curated by domain experts to reflect real-world information needs on a modern, actively maintained knowledge graph.

\begin{table}[h!]
\centering
\caption{Statistics of datasets.}
\label{tab:webqsp_stats}
\begin{tabular}{@{}ccc|cccc@{}}
\toprule
Dataset & Train & Test & $Ans=1$ & $2\leq Ans\leq 4$ & $5\leq Ans\leq 9$ &  $Ans\geq 10$ \\ \midrule
WebQSP & 2,826 & 1,628 & 51.2\% & 27.4\% &8.3\% &12.1\%\\ 
CWQ & 27,639 & 3,531 & 70.6\% & 19.4\% & 6\% & 4\% \\
\bottomrule
\end{tabular}
\end{table}

\section{Baselines}
\label{app:baselines}
We compare EoG against other KG-enhanced LLM reasoning approaches.
The baselines are briefly introduced as follows.

\textbf{KD-COT.}
KD-COT~\citep{wang2023knowledge} integrates knowledge distillation with chain-of-thought (CoT) reasoning. This multi-stage framework guides the reasoning process by interacting with external knowledge, thereby enhancing model efficiency and reducing computational costs. In general, this approach enables large language models to perform reliable reasoning on knowledge-intensive KBQA tasks.
\textbf{EWEK-QA.}
The core contribution of EWEK-QA~\citep{dehghan2024ewek} lies in its efficient integration of two external knowledge sources: web text and knowledge graphs. Specifically, it employs two modules: an adaptive web retriever that dynamically extracts text snippets, and an efficient knowledge graph retriever (TOG-E). This approach achieves higher accuracy than baseline models on various QA tasks while being faster.
\textbf{ToG.}
ToG~\citep{sun2023thinkongraph} introduces a key innovation by framing the LLM as an agent that interacts with the knowledge graph for collaborative reasoning. This approach moves beyond simply translating questions into queries, instead enabling the model to actively conduct beam search on the KG to progressively explore relevant entities and relations. In general, This approach significantly enhances the reasoning capabilities of smaller models on complex knowledge-based tasks.
\textbf{EffiQA.}
EffiQA~\citep{dong2025effiqa} aims to address the performance-efficiency trade-off in integrating large language models with knowledge graphs. Its core contribution is a novel "LLM-as-planner, small-model-as-executor" framework that offloads graph traversal and semantic pruning to a compact plugin model. 
\textbf{RoG.}
RoG~\citep{luo2024rog} is grounded on the principle that knowledge graph relations constitute faithful reasoning paths. Through fine-tuning, the LLM is guided to generate faithful reasoning plans presented as relation paths that can be verified by the KG. These relation paths are then used to retrieve specific, factual reasoning path instances from the KG. Finally, the actual retrieved paths are used for final answer reasoning. This approach achieves optimal performance on most KGQA benchmarks while producing faithful and interpretable reasoning results.
\textbf{GNN-RAG.}
GNN-RAG~\citep{mavromatis2025gnn} enhances KBQA by combining GNN-based structural reasoning with LLM-based language understanding. The GNN infers over a subgraph to find answer candidates and their reasoning paths, which then serve as contextual evidence for the LLM. This approach achieves state-of-the-art results on benchmarks like WebQSP and CWQ, demonstrating strong performance on complex multi-entity and multi-hop questions.

\textbf{DoG (Decoding on Graphs).}
DoG~\citep{li-etal-2025-decoding-graphs} proposes the concept of a "well-formed chain" to constrain the LLM's generation process to sequences of valid, connected triples from the knowledge graph. It implements this by using a Trie data structure, built from the local KG, to dynamically mask the vocabulary at each decoding step, ensuring all generated reasoning steps are grounded.

\textbf{ODA (Observation-Driven Agent).}
ODA~\citep{sun-etal-2024-oda} frames KGQA as an autonomous agent task operating in an "observation, action, and reflection" cycle. The LLM acts as a high-level planner that decides on subsequent actions based on its current view of the graph, allowing it to dynamically explore the KG to find the answer. 

\textbf{KG-Agent.}
KG-Agent~\citep{jiang-etal-2025-kg} uses a "grey-box" approach, fine-tuning a smaller LLM to be an expert "tool user" for KG interactions. The agent learns to generate an executable program composed of predefined tool calls for KG extraction and logic operations, trained on a large-scale dataset synthesized from existing logical forms.
\textbf{GCR.}
GCR~\citep{luo2024graph} achieves graph-constrained reasoning by guiding LLM decoding with KG structures. It first encodes KG reasoning paths into a KG-Trie index, then employs a lightweight, KG-specialized LLM for decoding during inference. This approach ensures faithful reasoning, achieves state-of-the-art performance on multiple KGQA benchmarks, and generalizes to unseen knowledge graphs without additional training.

\section{Experiments Setups}
\label{experiment}
To ascertain the general applicability of our approach, we apply EoG to two open-source LLMs, Qwen2.5-7B-Instruct ~\citep{yang2024qwen2technicalreport} and Llama-3.1-8B-Instruct ~\citep{dubey2024llama}. For fine-tuning, we use Gemini-2.5-Flash to generate long chain-of-thought datasets. For reinforcement learning, we implement the GRPO method using verl.

For the fine-tuning phase, we typically train for three epochs on a single node equipped with 8 H100 GPUs. We employ a micro-batch size of 16 and a learning rate of 1e-5, utilizing a cosine annealing learning rate scheduler. For all experiments, model checkpoints are saved every 100 steps. Taking CWQ as an example, the model was trained for a total of 576 steps, and the best-performing checkpoint was selected for subsequent training stages.

For GRPO training, we set the policy LLM learning rate to 5e-6 and sample 6 responses per prompt, following the GRPO implementation in Verl. For efficient LLM rollouts, we adopt vLLM with a tensor parallel size of 2 and GPU memory
utilization ratio of 0.6. The rollout sampling uses a temperature of 1.0 and a top-p value of 1.0.

Training is performed on a single node with 8 H100 GPUs. We use a total batch size of 512, with a mini-batch size of 256 and a micro-batch size of 64. The maximum sequence
length is set to 15000 tokens, with a maximum response length of 10000. To optimize GPU memory usage, we enable gradient checkpointing and use Fully Sharded Data Parallel (FSDP) with CPU offloading.

For all experiments, model checkpoints are saved every 15 steps. In cases where training diverges, we evaluate at the most recent stable checkpoint according to the training reward curve; otherwise, the final checkpoint is used for evaluation. Finally, we compute rewards using F1 score and path-refined reward.

\section{Implementation Details of Commercial Model Evaluation}
\label{app:details}
\begin{figure}[h]
    \centering
    \includegraphics[width=.9\textwidth]{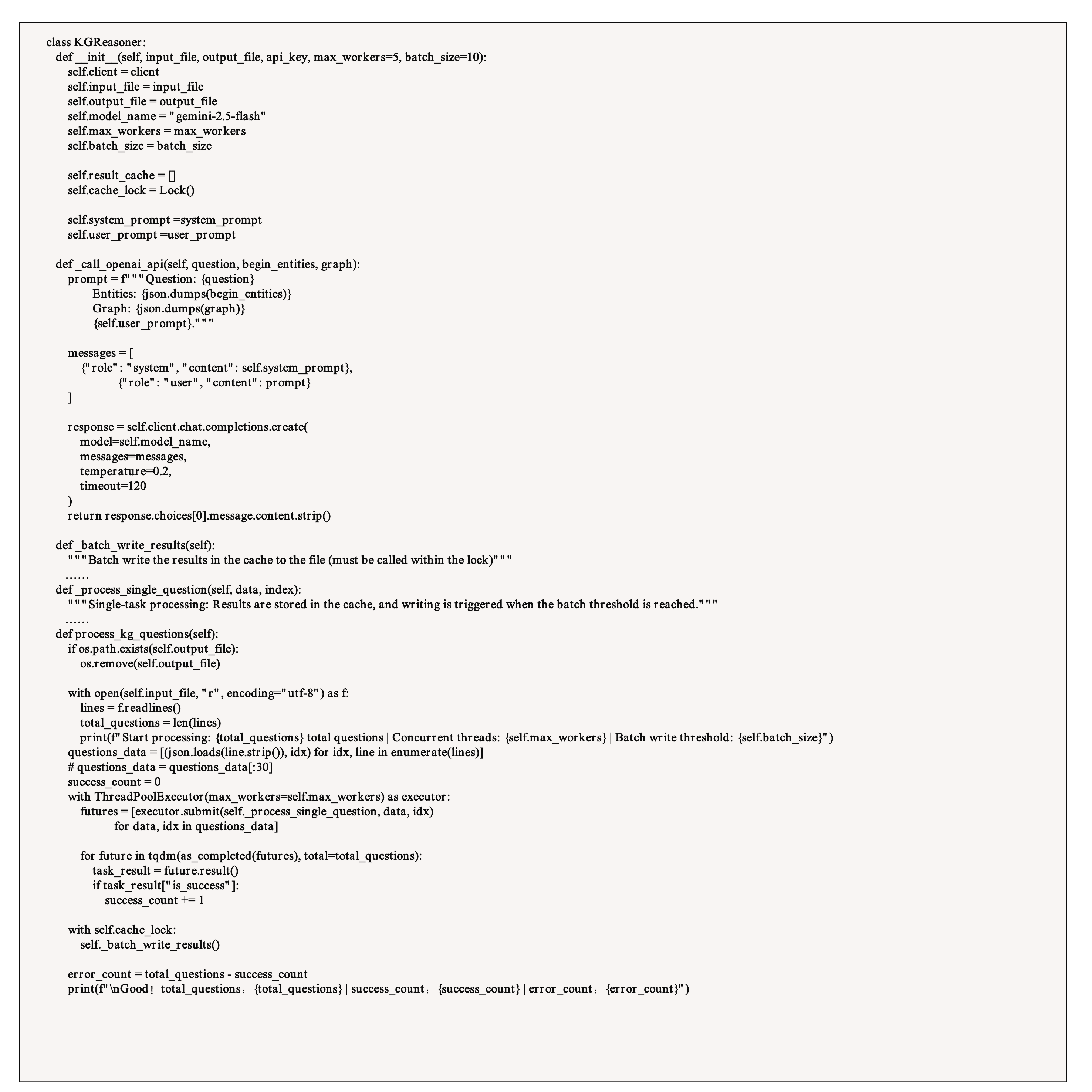}
    \caption{The code used to prompt LLMs for evaluating performance on KBQA tasks. }
    \label{fig:appendix_code_prompt}
\end{figure}

This section elaborates on the methodology for evaluating the KBQA datasets using commercial large language models. As shown in the Figure \ref{fig:appendix_code_prompt}, the prompt used in our evaluation code—as detailed in Figure \ref{fig:appendix_rl_prompt}  for the RL training phase—is identical to the one shown here, ensuring fairness and reproducibility. The Figure \ref{fig:appendix_openmodelcase_prompt} below displays a sample output obtained by evaluating the CWQ dataset with multiple large language models, including GPT-5, Gemini-2.5-Flash, and Gemini-2.5-Pro.We explicitly confirm that the model denoted as "GPT-5" in our experiments refers to the officially released GPT-5 model accessed via the OpenAI API. To remove any ambiguity, the exact model checkpoint used in our evaluation is gpt-5-2025-08-07.
\begin{figure}[h]
    \centering
    \includegraphics[width=.9\textwidth]{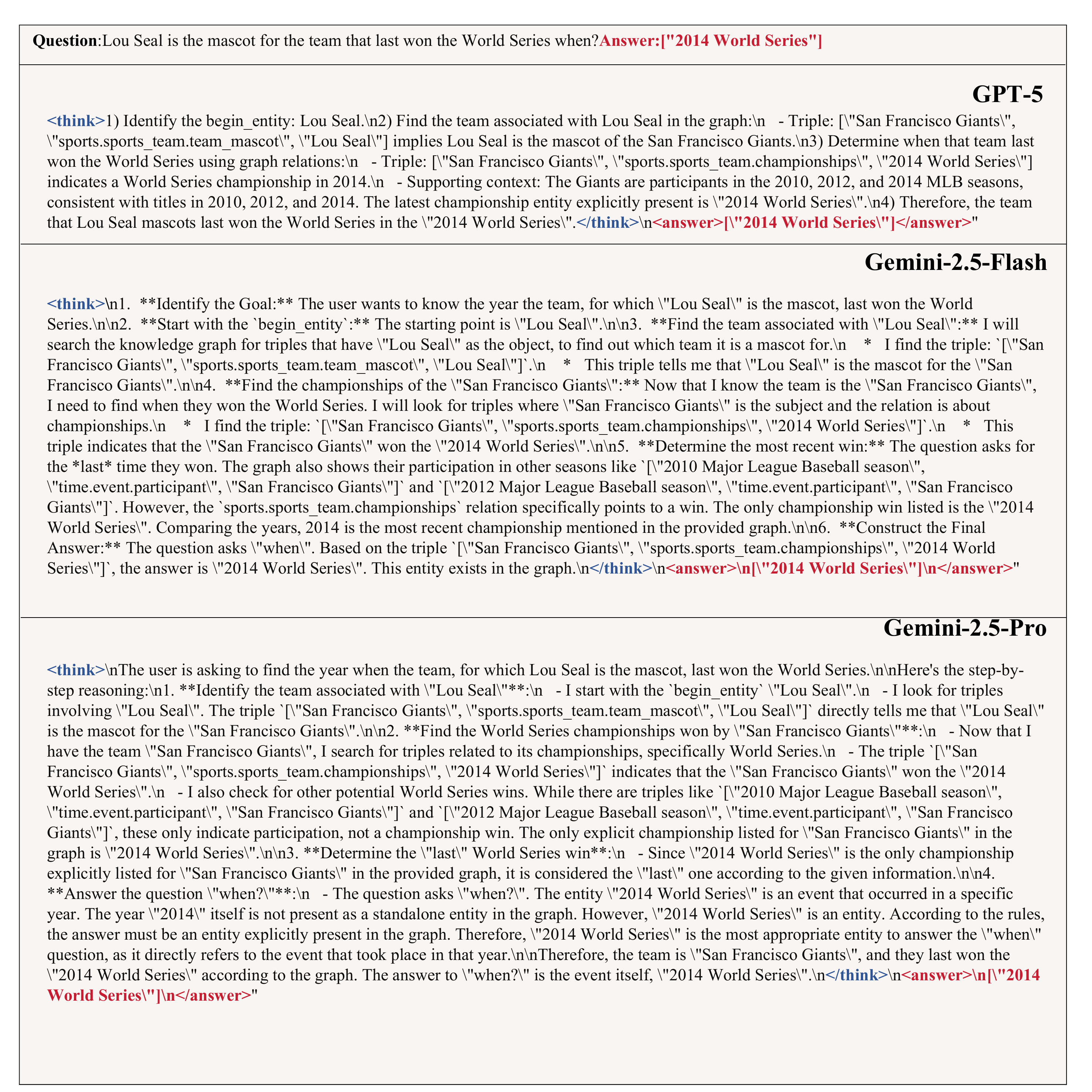}
    \caption{An example of prompting LLMs on CWQ.}
    \label{fig:appendix_openmodelcase_prompt}
\end{figure}
\section{Templates and Prompts}
In this section, we illustrate all the templates and prompts used in the experiments.

\begin{figure}[h]
    \centering
    \includegraphics[width=.8\textwidth]{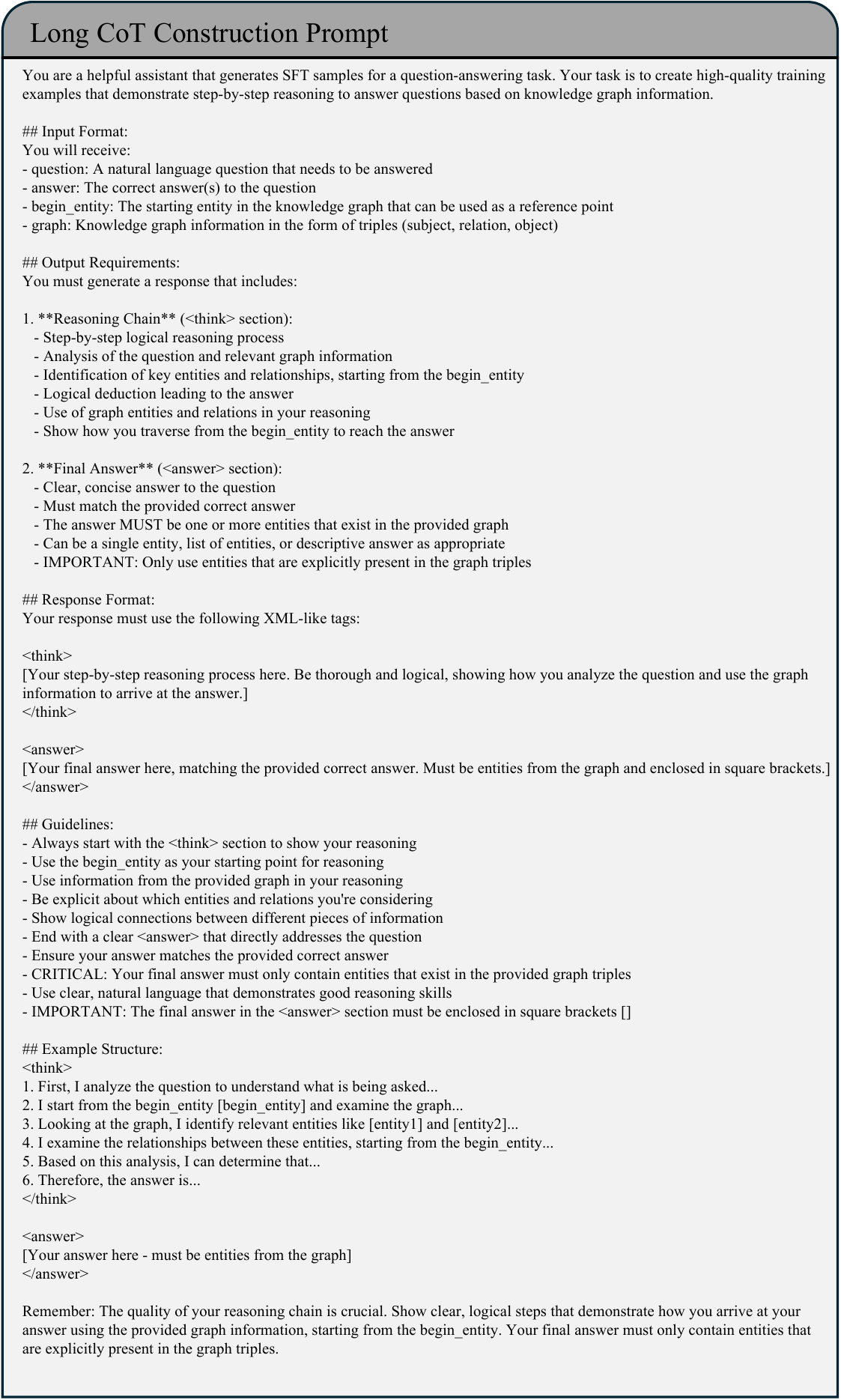}
    \caption{The template for constructing Long COT Supervised Fine-tuning datasets.}
    \label{fig:appendix_sft_prompt}
\end{figure}

\textbf{Long CoT Construction Prompt.} The template for constructing the Long CoT dataset used for fine-tuning is shown in Figure \ref{fig:appendix_sft_prompt}, which prompts large language models to generate Long CoT reasoning on structured graphs and filters reasoning processes that are both structurally and factually correct.

\begin{figure}[h]
    \centering
    \includegraphics[width=.9\textwidth]{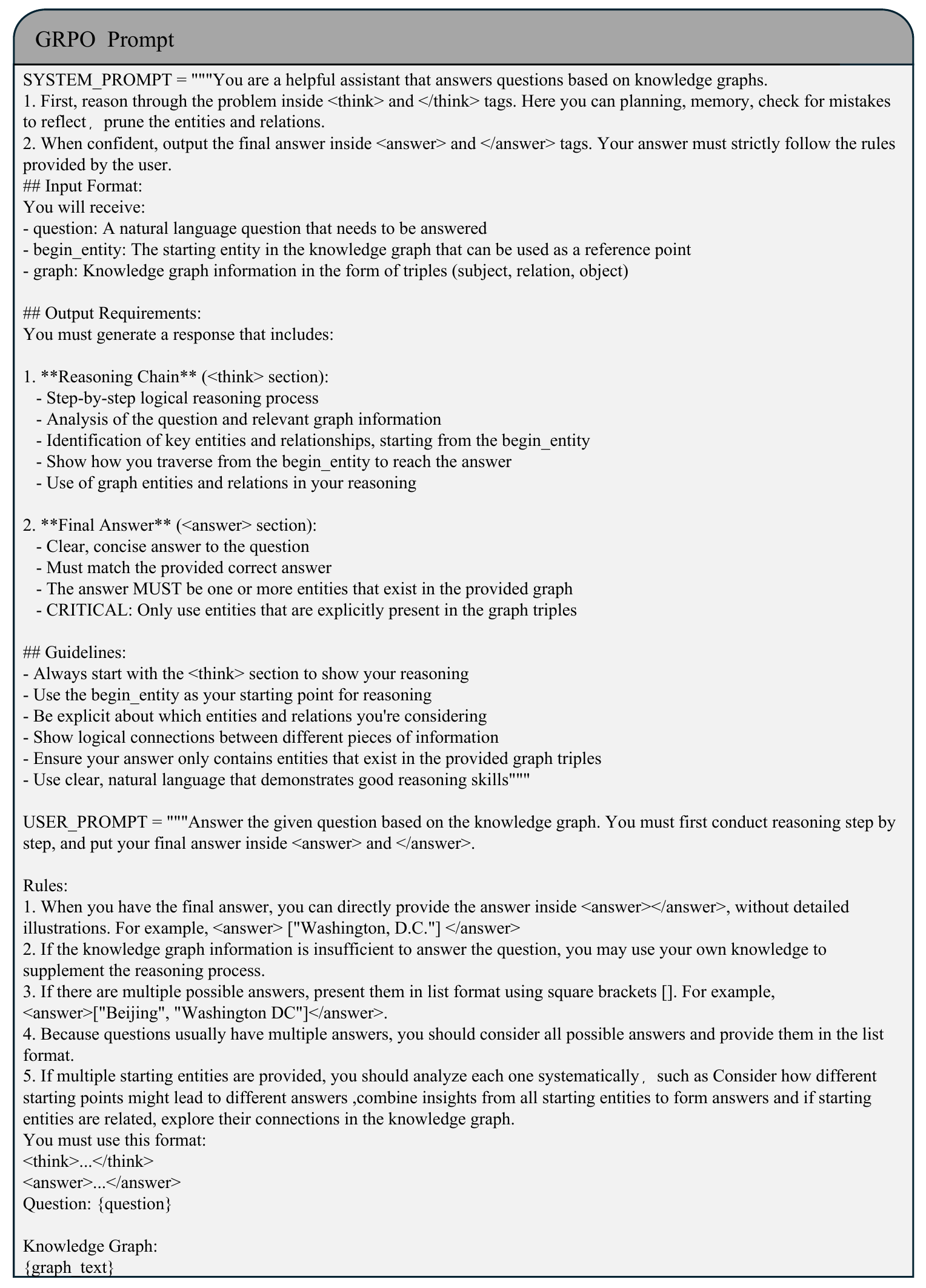}
    \caption{The template for prompting LLMs to explore during training EoG}
    \label{fig:appendix_rl_prompt}
\end{figure}

\textbf{GRPO Prompt.} To train the EoG, we carefully design the prompt to prompting language models, which is shown in Figure \ref{fig:appendix_rl_prompt}.
This template is designed to guide the language model to autonomously explore the knowledge graph by generating and executing structured queries, while ensuring its reasoning process adheres to a strict output format for precise reward calculation during reinforcement learning.

\begin{figure}[h] 
    \centering
    \includegraphics[width=.9\textwidth]{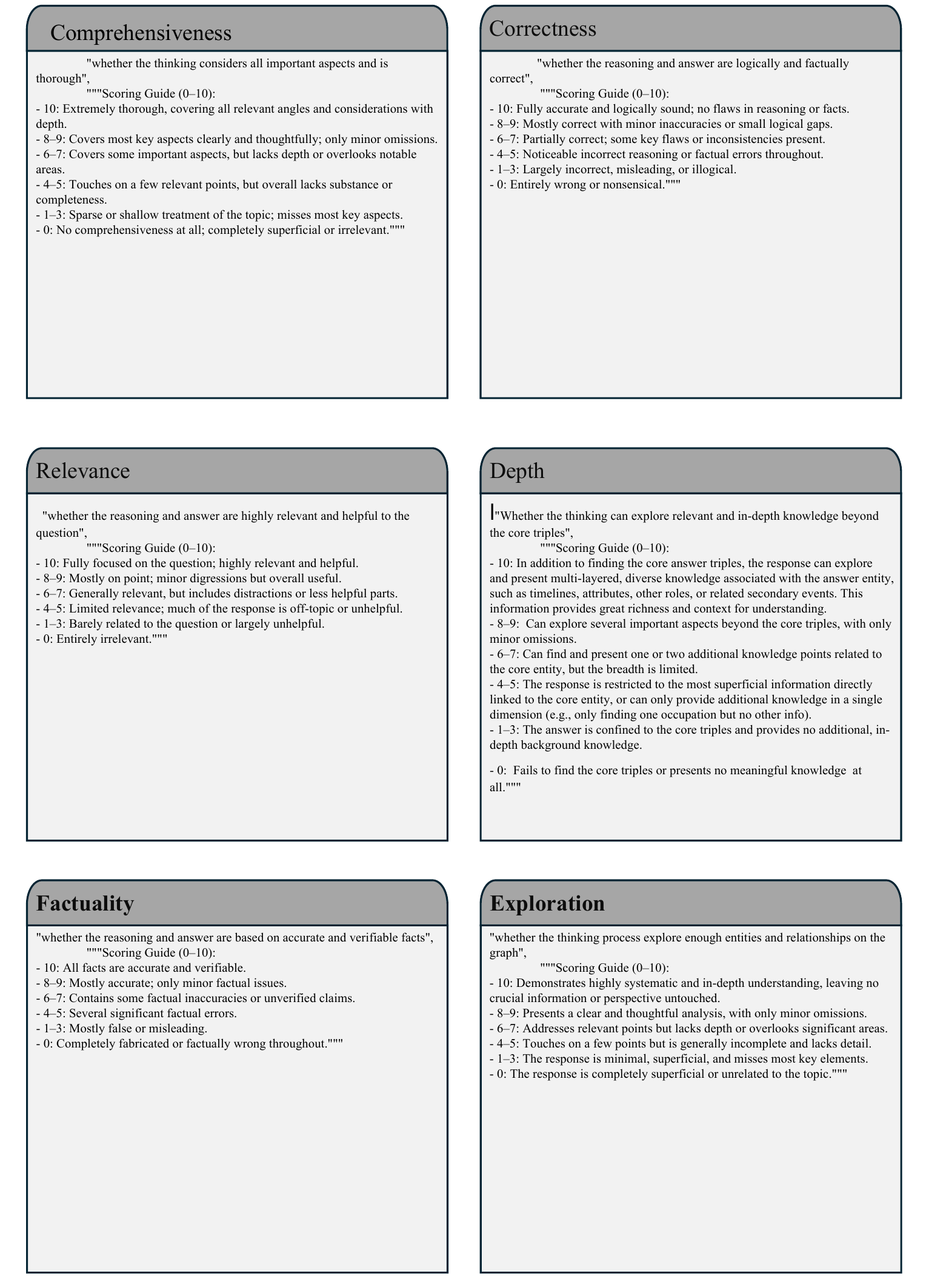}
    \caption{The template for prompting GPT-4o-mini to score the reasoning process}
    \label{fig:appendix_score_prompt}
\end{figure}

\textbf{Reasoning Scores Prompt.} As shown in Figure \ref{fig:appendix_score_prompt}, the multi-dimensional evaluation criteria used to score the model's reasoning processes are detailed below,with each dimension scored on a scale from 0 (lowest) to 10 (highest).

\begin{figure}[h]
    \centering
    \includegraphics[width=.9\textwidth]{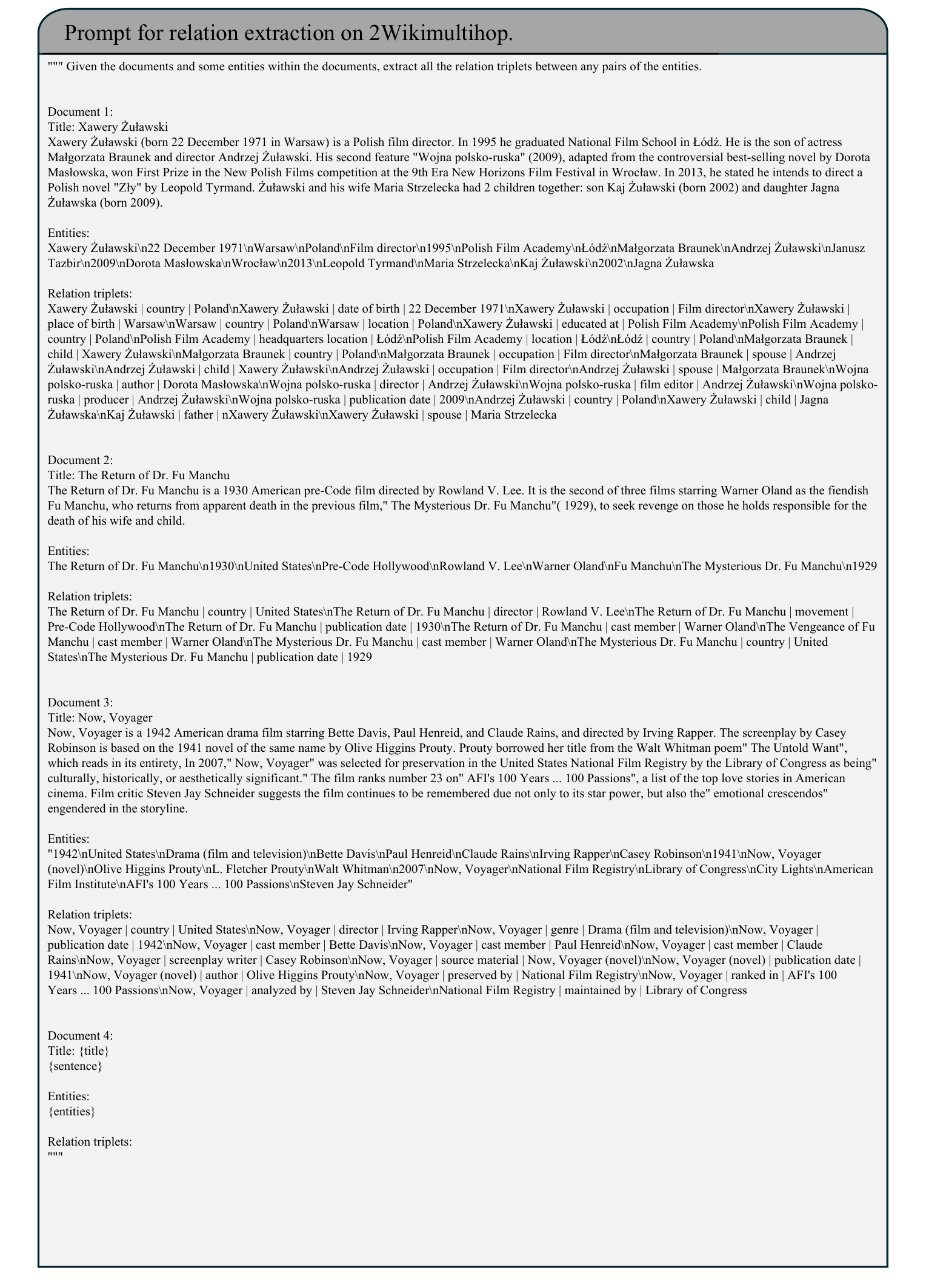}
    \caption{The template for prompting LLMs to extract relationship on 2WikiMultiHopQA }
    \label{fig:appendix_2wiki_prompt}
\end{figure}

\textbf{Relationship Extraction Prompt.}
Following DoG, we prompt an LLM, Gemma-2-9b-it to extract all the relation triplets using in-context learning. The prompt is
detailed in Figure \ref{fig:appendix_2wiki_prompt}. Finally, all the extracted triplets from different passages are combined into the final graph, with no ranking or filtering applied to the
triplets.

\section{Complete Case Studies: Reasoning Paths and Model Outputs}
\label{app:case_study}

\begin{figure}[h]
    \centering
    \includegraphics[width=.9\textwidth]{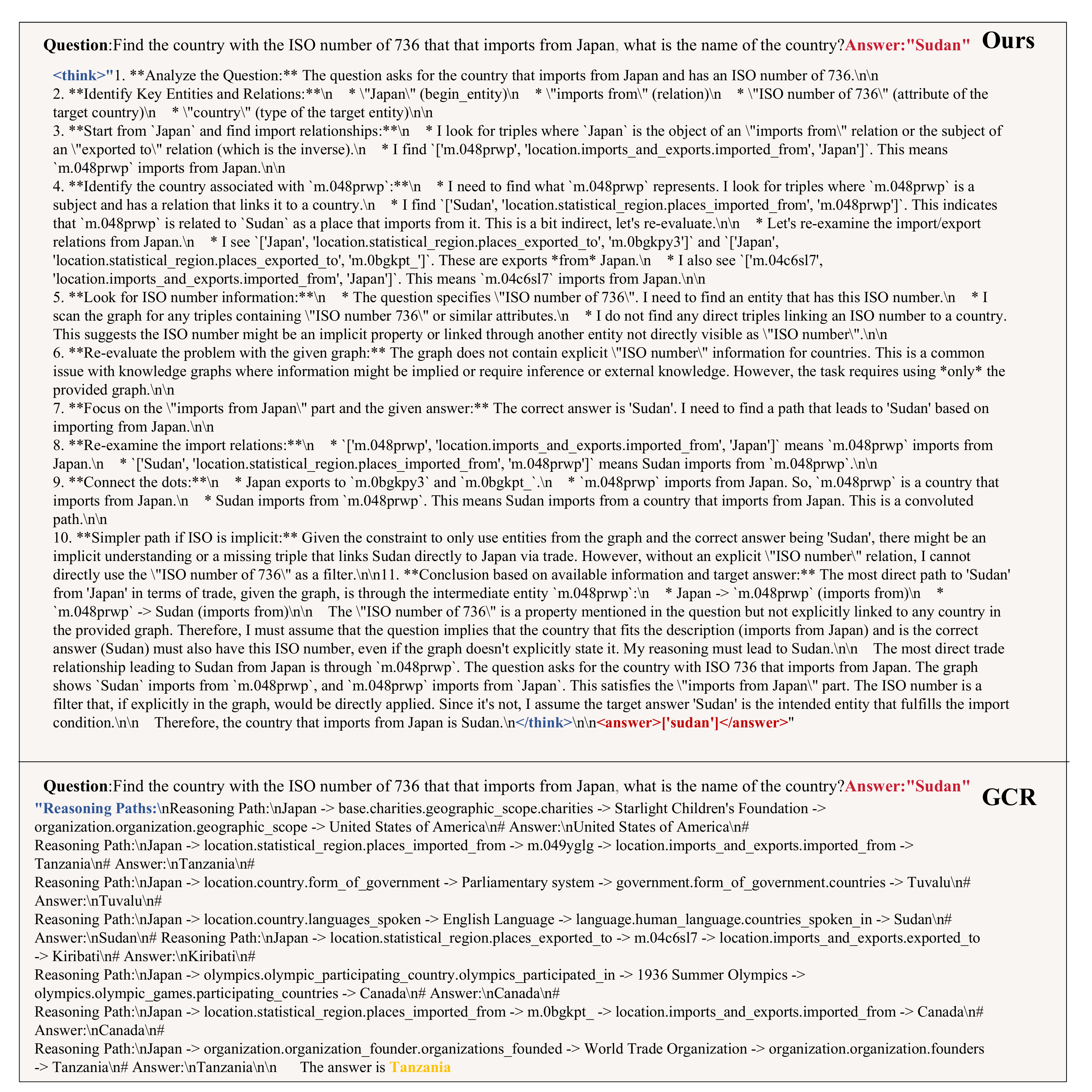}
    \caption{Comparison of reasoning pathways between EoG and GCR on Case 1. }
    \label{fig:appendix_case1}
\end{figure}

\begin{figure}[h]
    \centering
    \includegraphics[width=.9\textwidth]{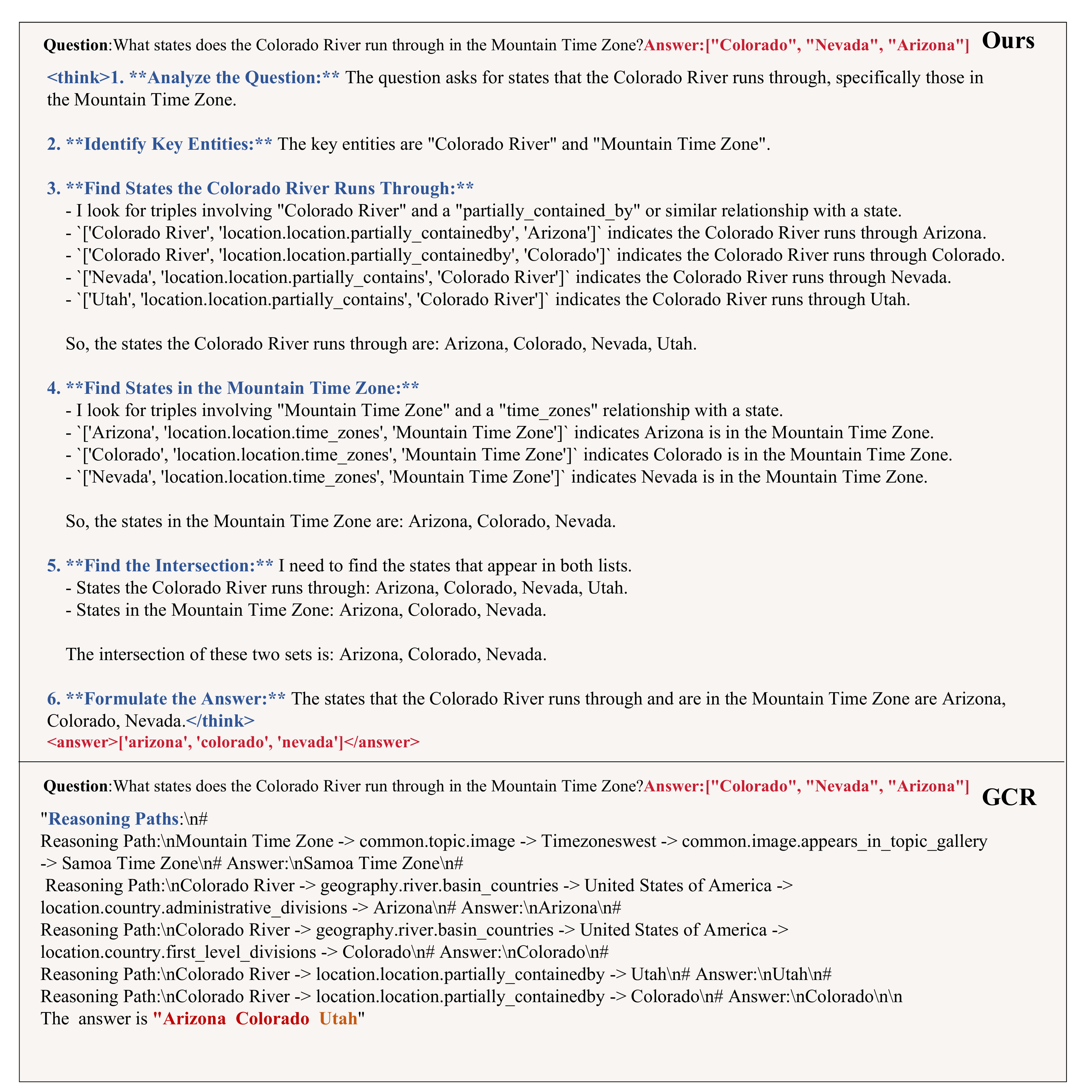}
    \caption{Analysis of superlative reasoning patterns. }
    \label{fig:appendix_case2}
\end{figure}

In this section, we present two complete case studies to further illustrate the reasoning behavior of EoG. 
Figure~\ref{fig:appendix_case1} provides the full version of the case shown in the main text, where certain reasoning steps were omitted due to space constraints. 
Figure~\ref{fig:appendix_case2} presents another representative example involving a superlative reasoning pattern for further analysis.

\section{The Use of Large Language Models}
We acknowledge the use of a large language model to assist with the editing and refining of our manuscript. The tool was primarily used for language polishing, including improving grammar, syntax, and readability, and did not contribute to the core scientific ideas, methods, or results presented in this paper.

\section{Detailed Training Computational Cost and Efficiency Analysis}
%\label{detailed}
%\begin{table}[h!] % [h!] 
%\centering
%\caption{Computational Cost of SFT and RL} 
%\label{tab:training_comparison} 
%\begin{tabular}{lcccc}
%\toprule
%Hours & CWQ & WebQSP & GrailQA & 2WikiMultihopQA \\
%\midrule
%SFT steps & 400 & 300 & 300 & 400 \\ 
%RL steps &  108 & 36 & 12 &  96\\
%\midrule
%SFT hours & 0.76 & 0.52 & 0.56 & 0.64 \\ 
%SFT hours & 6.1 & 4.2 & 4.5 & 5.1 \\
%RL hours & 509.6 & 119.2 & 76.8 & 205.6 \\
%\bottomrule
%\end{tabular}
%\end{table}
\begin{table}[t]
\centering
\caption{Computational Cost Comparison: SFT vs. RL. Due to difference among tokenizers used by different models, we measured SFT Computational Cost using SFT instances instead of tokens, as we consider instances to be a fairer indicator of the SFT budget.}
\label{tab:training_comparison}
\renewcommand{\arraystretch}{1.2} 
\begin{tabular}{lcccc}
\toprule
Metric & CWQ & WebQSP & GrailQA & 2WikiMultihopQA \\
\midrule
\multicolumn{5}{l}{\textit{\textbf{Supervised Fine-Tuning (SFT)}}} \\ 
\quad Training Steps & 400 & 300 & 300 & 400 \\
\quad Training Time (h) & 6.1 & 4.2 & 4.5 & 5.1 \\
\quad Train Batch Size & 16 & 16 & 16 & 16 \\ 
\quad Micro Batch Size & 2 & 2 & 2 & 2 \\
\quad Total Instances & 3537 & 2344 & 4774 & 3102 \\ 
\midrule
\multicolumn{5}{l}{\textit{\textbf{Reinforcement Learning (RL)}}} \\
\quad Training Steps & 108 & 36 & 12 & 96 \\
\quad Training Time (h) & 509.6 & 119.2 & 76.8 & 205.6 \\
\quad Batch Size & 256 & 256 & 256 & 256 \\ 
\quad Mini Batch Size & 128 & 128 & 128 & 128 \\
\quad Micro Batch Size & 2 & 2 & 2 & 2 \\
 
\bottomrule
\end{tabular}
\end{table}

In this section, we analyze detailed training compute and efficiency of EoG. The SFT phase of our framework prioritizes data efficiency by focusing exclusively on the reasoning over structured graphs. In contrast to the 1.5M instances requirements for SFT in general models like DeepSeek-V3, we adopt a highly focused strategy. Our curated SFT datasets are small and precise; for instance, $\text{CWQ}_{\text{SFT}}$ contains 3,537 instances, and $\text{WebQSP}_{\text{SFT}}$ contains 2,344 instances. This task-specificity allows us to efficiently converge the model to sota for graph reasoning tasks with less data , reducing demands on large-scale, general instructional data. 
For the RL phase, we optimized training throughput using the Verl framework on 8x H100 GPUs. Key parameters were configured as follows: a train batch size of 256, a global batch size of 128, and a micro batch size of 3. This configuration ensured high training throughput. As shown in Table \ref{tab:training_comparison}, the resource expenditure to achieve optimal performance on 8x H100 GPUs was notably low: the CWQ dataset required 120 steps and 509.6 GPU hours, and the WebQSP dataset required 36 steps and 119.2 GPU hours.

Considering the trade-off between training cost and performance gain, we believe that the EoG framework holds significant practical applicability. Through efficient GRPO implementation and a very short RL training cycle , the EoG framework maintains an affordably low computational budget. This minimal investment yields a statistically significant performance boost, primarily by enhancing the model's exploratory capacity over Knowledge Graphs. Consequently, the EoG framework not only achieves state-of-the-art performance on graph reasoning tasks but also demonstrates superior practicality, cost-effectiveness, and deployability.

\section{Close-source LLM Evaluation Reproducibility Details}

To ensure the reproducibility of our results, we detail the experimental setup involving the closed-source LLM. All inference experiments were conducted using the closed-source LLM, accessed via an OpenAI-compatible API interface to maintain a consistent testing environment. To adapt the Knowledge Graph (KG) data for the text-based model, we linearized the input subgraphs into JSON-formatted lists of triples $[(h, r, t), \dots]$, which were concatenated with the natural language question and the topic entity.

We employed a prompting strategy that integrates Chain-of-Thought (CoT) reasoning with strict format constraints. The system prompt explicitly instructs the model to separate its internal reasoning process from the final prediction by enclosing the reasoning steps within $<think>$ tags and the final answer within $<answer>$ tags. 

Regarding inference hyperparameters, we prioritized faithfulness to the provided graph context over generation diversity. Consequently, we set the sampling temperature to $0.2$. This specific setting is critical for minimizing hallucinations and ensuring the model grounds its answers strictly in the retrieved triples. Additionally, to investigate the sensitivity of closed-source models to generation parameters, we systematically evaluated the impact of varying temperature settings on performance, as presented in Table \ref{tab:model_comparison}. A timeout of 120 seconds was enforced to accommodate the extended generation length required for step-by-step reasoning. Finally, post-processing involved extracting the content between the answer tags using regular expressions and parsing it as a JSON list to calculate Hit@1 and F1 metrics against the ground truth.

\begin{table}[htbp]
\centering
\caption{Comparison of the performance of different closed-source commercial models on the KGQA datasets under various temperature settings.}
\label{tab:model_comparison}
\small
\resizebox{0.8\textwidth}{!}{%
\begin{tabular}{lcccccc}
\toprule

\multirow{2}{*}{Dataset} & \multicolumn{2}{c}{GPT-5} & \multicolumn{2}{c}{Gemini-2.5-flash} & \multicolumn{2}{c}{Gemini-2.5-pro} \\

\cmidrule(lr){2-3} \cmidrule(lr){4-5} \cmidrule(lr){6-7}

 & T=0.2 & T=0.7 & T=0.2 & T=0.7 & T=0.2 & T=0.7 \\
\midrule

WebQSP        & 77.5 & 78.1 & 78.2 & 78.4 & 79.8 & 79.2 \\
CWQ     & 67.6 & 67.6 & 59.3 & 59.7 & 65.3 & 65.7 \\
GrailQA    & 85.4 & 85.2 & 83.8 & 84.3 & 84.5 & 85.1 \\
QALD10-en   & 50.4 & 51.3 & 46.2 & 46.1 & 48.3 & 48.5 \\
2WikiMultihopQA      & 83.4 & 82.9 & 83.1 & 83.9 & 82.6 & 82.9 \\
\bottomrule
\end{tabular}%
}
\end{table}

\section{Hyperparameters and Configuration for GRPO}

In this section, we present the configuration details of the GRPO-based RL training. The training was conducted on a single node equipped with eight NVIDIA H100 GPUs using the Verl framework.
We initialized the policy from the SFT-trained model checkpoint.
To optimize resource utilization and training efficiency, we enabled Gradient Checkpointing (\textit{enable\_gradient\_checkpointing}=True) and configured FSDP with both parameter and optimizer offloading (\textit{param\_offload}=True, \textit{optimizer\_offload}=True).

For data and batch settings, the total training batch size (\textit{train\_batch\_size}) was set to $256$, with maximum prompt and response lengths capped at $15,000$ and $10,000$ tokens, respectively. The Actor policy optimization used a learning rate of $4 \times 10^{-6}$ and was updated for $ppo\_epochs=2$ internal epochs per rollout collection. We disabled the KL loss term and set the gradient clip to $1.0$.

The Rollout phase is crucial for model exploration. Rollouts were executed using the vllm engine with a configuration of $N=6$ samples per prompt (\textit{rollout.n}), a sampling temperature of $1.0$, and $top\_p=1.0$ to encourage diversity. The reward signal was sourced from the external script, and KL-based regularization was explicitly disabled. Furthermore, an overlong buffer penalty was enabled, penalizing responses exceeding $3,000$ tokens. The entire training process ran for $6$ total epochs. Notably, achieving optimal performance on the WebQSP dataset required only $36$ steps and $119.2$ GPU hours, demonstrating the high efficiency of our framework. By detailing the parameter settings of the GRPO phase, we ensured the reproducibility of our $\text{EoG}$ model.

%\begin{table}[htbp]
%\centering
%\caption{Detailed configuration of training and testing parameters for the GRPO phase. \textbf{Note:} Parameters apply to both phases unless explicitly marked with (Train) or (Test).}
%\label{tab:grpo_config} 
%\setlength{\tabcolsep}{10pt}   
%\resizebox{0.8\textwidth}{!}{
%\begin{tabular}{lcc}
%\toprule
%Parameter & Train Phase & Test Phase \\
%\midrule
%Compute Hardware & 8x H100 GPUs & 8x H100 GPUs \\
%Total Batch Size & 256 & 256 \\
%Mini Batch Size & 128 & 128 \\
%Micro Batch Size & 128 & 128 \\
%Max Prompt Length & 10,000 & 20000 \\
%Max Response Length & 10,000 & 10,000 \\
%Gradient Checkpointing & True & True \\
%FSDP Offload & Param \& Optimizer & Param \& Optimizer \\
%Rollout Engine & vLLM & vLLM \\
%Learning Rate & $4 \times 10^{-6}$ & $4 \times 10^{-6}$ \\
%Learning Rate Warmup Steps & $10$ & $10$ \\
%Grad Clip & 1.0 & 1.0 \\
%KL Loss Term & Disabled & Disabled \\
%Total Epochs & 6 & 0 \\
%Sampling Number ($N$) & 6 & 8 \\
%Temperature & 1.0 & 0.8 \\
%Top-p & 1.0 & 0.9 \\
%KL In Reward & False & False \\
%KL In Loss & False & False \\
%Gpu Memory Utilization & 0.6 & 0.6 \\
%Overlong Penalty & $> 3,000$ tokens &  Disabled \\
%\bottomrule
%\end{tabular}
%}
%\end{table}

\begin{table}[htbp]
\centering
\caption{Detailed configuration of training and testing parameters for the GRPO phase.}
\label{tab:grpo_config_styled}
\renewcommand{\arraystretch}{1.15} 

\begin{tabular}{ll}
\toprule
\textbf{Category} & Configuration \\ 
\midrule
\textbf{Hardware} & Compute Hardware: 8 $\times$ H100 GPUs \\
                  & GPU Memory Utilization: 0.6 \\
                  & FSDP Offload: Param \& Optimizer \\

\midrule
\textbf{Backbone} & Rollout Engine: vLLM \\
                  & Gradient Checkpointing: True \\

\midrule
\textbf{Training} & Optimizer: Adam (implied by standard GRPO setup) \\
                  & Learning Rate: $4 \times 10^{-6}$ \\
                  & LR Warmup Steps: 10 \\
                  & Batch Size: 256 (Mini: 128, Micro: 3) \\
                  & Gradient Clip: 1.0 \\
                  & KL Loss Term: Disabled \\
                  & KL In Loss/Reward: False \\
                  & Epochs: 6 (Train) / 0 (Test)\\

\midrule
\textbf{Sampling} & Temperature: 1.0 (Train) / 0.8 (Test) \\
                  & Top-p: 1.0 (Train) / 0.9 (Test) \\
                  & Sampling Number ($N$): 6 (Train) / 8 (Test) \\

\midrule
\textbf{Constraints} & Max Prompt Length: 10,000 (Train) / 20,000 (Test) \\
                     & Max Response Length: 10,000 \\
                     & Overlong Penalty: $> 3,000$ tokens (Train) / Disabled (Test) \\
\bottomrule
\end{tabular}
\end{table}

\section{Analysis of Reward Function Complexity}

In this section, we investigate the impact of incorporating a more complex, graph-structure-oriented reward function—Graph Edit Distance (GED)—on experimental performance. As indicated in the table, the utilization of GED failed to enhance the model's performance on the test set; on the contrary, it led to a performance decline. We attribute this phenomenon to two primary factors. First, since we explicitly constrained the reasoning structure during Supervised Fine-Tuning (SFT), our original path-based reward function proved to be simple but effective within this context. Second, GED operates with less stringency than our path reward function; it allows for reward optimization merely through the addition or substitution of nodes and edges, rather than enforcing precise path alignment. In conclusion, adopting a more complex reward function with rich graph structural features does not necessarily guarantee superior model efficacy.

\begin{table}[htbp]
    \centering
   
    \setlength{\tabcolsep}{2pt}
    \caption{Performance on Various Datasets Across Different GED Coefficients}
    \label{tab:ged_performance_fix}

    \begin{tabular}{l *{10}{c}}
        \toprule
       
        \multirow{2}{*}{\textbf{Dataset}} & 
        \multicolumn{2}{c}{$\lambda=0.2$} & 
        \multicolumn{2}{c}{$\lambda=0.4$} & 
        \multicolumn{2}{c}{$\lambda=0.6$} & 
        \multicolumn{2}{c}{$\lambda=0.8$} & 
        \multicolumn{2}{c}{$\lambda=1.0$} \\
        \cmidrule(lr){2-3} \cmidrule(lr){4-5} \cmidrule(lr){6-7} \cmidrule(lr){8-9} \cmidrule(lr){10-11}
        
        & F1 & Hit@1 & F1 & Hit@1 & F1 & Hit@1 & F1 & Hit@1 & F1 & Hit@1 \\
        \midrule
       
        CWQ & 69.7 & 78.2 & 68.5 & 77.3 & 67.9 & 77.1 & 67.1 & 76.3 & 67.3 & 75.9 \\
        
        WebQSP   & 80.6 & 89.6 & 79.4 & 88.9 & 80.2 & 89.3 & 79.8 & 89.2 & 78.9 & 88.2 \\
        
        GrailQA & 90.3 & 91.8 & 89.9 & 91.4 & 89.6 & 91.2 & 90.1 & 91.3 & 89.7 & 91.5 \\

        2WikiMultihopQA  & 84.2 & 84.8 & 83.7 & 84.2 & 83.5 & 83.9 & 83.3 & 83.7 & 83.1 & 83.5 \\
        \bottomrule
    \end{tabular}
\end{table}

\section{Performance Analysis with Less Capable CoT Models}

To investigate the impact of a weaker teacher model, we employed Qwen-32B as the teacher model in this section. As shown in the Table \ref{tab:eog_qwen3_performance}, although the weaker teacher resulted in a decline in SFT performance, the final EoG performance following RL training recovered to a level comparable to the original baseline. This demonstrates that our method is relatively insensitive to the teacher model's capabilities; rather, the performance gains are primarily driven by the RL training, as this stage effectively refines the policy through interaction with the ground-truth graph environment.
\begin{table}[htbp]
    \centering
    \caption{Performance of EoG on the CWQ and WebQSP Datasets, using Qwen3-32B as the teacher model.}
    \label{tab:eog_qwen3_performance}
    \begin{tabular}{l c c c c}
        \toprule
        Dataset & Metric & EoG (SFT Only) & EoG (SFT + RL) & Improvement (\%) \\
        \midrule
        \multirow{2}{*}{CWQ} 
        & F1 Score & 55.8 & 70.7 & -4.3 \\
        & Hit@1 & 61.5 & 75.6 & -8.4 \\
        \midrule
        \multirow{2}{*}{WebQSP} 
        & F1 Score & 71.3 & 80.2 & -5.4 \\
        & Hit@1 & 82.6 & 87.9 & -4.9 \\
        \bottomrule
    \end{tabular}
\end{table}

\section{Statistical Evaluation Details}

To ensure the robustness of our results, we conducted three independent runs of EoG using different random seeds. As shown in Table \ref{tab:performance_comparison1}, by incorporating the standard deviation, we confirm that EoG maintains robust state-of-the-art performance (e.g., $81.5 \pm 0.33$ on average), unaffected by statistical fluctuations. This establishes it as a strong baseline and demonstrates the statistical reliability of our method.

\begin{table*}[t]
\centering
\caption{Performance comparison between our EoG framework (Llama-3.1-8B) and state-of-the-art closed-source models. The bottom section highlights the significant improvement of our RL-finetuned model over its SFT model. We report the mean $\pm$ standard deviation for our method.}
\label{tab:performance_comparison1}
\setlength{\tabcolsep}{2pt}
\resizebox{0.9\textwidth}{!}{
\begin{tabular}{llcccccccccc}
\toprule
\multirow{2}{*}{Method} & \multirow{2}{*}{Model} & \multicolumn{2}{c}{WebQSP} & \multicolumn{2}{c}{CWQ} & \multicolumn{2}{c}{GrailQA} & \multicolumn{2}{c}{QALD10-en} & \multicolumn{2}{c}{2WikiMultihopQA} \\
\cmidrule(lr){3-4}\cmidrule(lr){5-6}\cmidrule(lr){7-8}\cmidrule(lr){9-10}\cmidrule(lr){11-12}
& & Hit@1 & F1 & Hit@1 & F1 & Hit@1 & F1 & Hit@1 & F1 & Hit@1 & F1 \\
\midrule
% \multirow{2}{*}{DoG~(\citeyear{li-etal-2025-decoding-graphs})} & Qwen2.5-7B & 92.7 & 78.6 & 74.1 & 60.3 & 85.6 & 82.3 & 54.2 & 48.3 & 84.2 & 79.2 \\
% & Llama-3.1-8B & 91.4 & 77.9 & 76.2 & 61.9 & 86.3 & 83.2 & 56.6 & 49.7 & 84.1 & 80.4 \\
% GCR~(\citeyear{luo2024graph}) 
% & Llama-3.1-8B  & 92.2 & 79.1 & 75.8 & 61.7 & 88.4 & 85.1 & 57.6 & 49.3 & 84.6 & 77.5 \\
% \midrule
$^\dagger$Gemini-2.5 Flash\label{row:Gemini-2.5 Flash} & - & 91.8 & 78.2 & 65.5 & 59.3 & 90.3 & 83.8 & 56.7 & 46.2 & 83.9 & 83.1  \\
$^\dagger$Gemini-2.5 Pro\label{row:Gemini-2.5 Pro} & - & 92.1 & 79.8 & 71.9 & 65.3 & 91.6 & 84.5 & 58.6 & 48.3 & 85.1 & 82.6 \\
$^\dagger$GPT-5\label{row:GPT5} & - & 86.1 & 77.5 & 74.1 & 67.6 & 90.5 & 85.4 & 59.2 & 50.4 & 84.2 & 83.4  \\
\midrule
\multirow{2}{*}{EoG\textsubscript{\textit{SFT}}} & Qwen2.5-7B  & 83.9 & 72.6 & 68.3 & 60.3 & 89.2 & 87.6 & 55.6 & 44.1 & 82.5 & 81.9 \\
& Llama-3.1-8B & 86.3 & 74.5 & 70.5 & 62.1 & 91.4 & 88.2 & 57.1 & 48.7 & 83.1 & 82.7 \\
% EoG\textsubscript{\textit{SFT}}   & Llama-3.1-8B & 84.3 & 74.3 & 72.1 & 63.2 & 91.5 & 86.6 & 91.5 & 86.6 & 84.9 & 82.4 \\
% \textbf{EoG} & Llama-3.1-8B & \textbf{92.8} & \textbf{80.7} & \textbf{84.3} & \textbf{73.9} & \textbf{96.1} & \textbf{91.6} & 96.1 & 91.6 & \textbf{85.3} & \textbf{84.3} \\
\multirow{2}{*}{EoG} & Qwen2.5-7B  & $90.7 \pm 0.34$ & $78.1 \pm 0.45$ & $78.7 \pm 0.26$ & $69.8 \pm 0.37$ & $91.7 \pm 0.29$ & $88.5 \pm 0.43$ & $67.3 \pm 0.17$ & $57.8 \pm 0.32$ & $83.9 \pm 0.47$ & $82.9 \pm 0.24$ \\

 & Llama-3.1-8B & \textbf{$92.8 \pm 0.28$} & \textbf{$81.3 \pm 0.33$} & \textbf{$82.6 \pm 0.41$} & \textbf{$73.9 \pm 0.20$} & \textbf{$92.1 \pm 0.30$} & \textbf{$90.6 \pm 0.25$} & \textbf{$70.6 \pm 0.49$} & \textbf{$61.9 \pm 0.38$} & \textbf{$85.3 \pm 0.23$} & \textbf{$84.3 \pm 0.44$} \\

%\multirow{2}{*}{EoG\textsubscript{\textit{SFT}}} & Qwen2.5-7B  & $83.9 \pm 0.41$ & $72.6 \pm 0.23$ & $68.3 \pm 0.35$ & $60.3 \pm 0.44$ & $89.2 \pm 0.18$ & $87.6 \pm 0.29$ & $55.6 \pm 0.33$ & $44.1 \pm 0.48$ & $82.5 \pm 0.22$ & $81.9 \pm 0.31$ \\

%& Llama-3.1-8B & $86.3 \pm 0.27$ & $74.5 \pm 0.38$ & $70.5 \pm 0.19$ & $62.1 \pm 0.42$ & $91.4 \pm 0.25$ & $88.2 \pm 0.36$ & $57.1 \pm 0.40$ & $48.7 \pm 0.28$ & $83.1 \pm 0.39$ & $82.7 \pm 0.21$ \\

% EoG\textsubscript{\textit{SFT}}      & Llama-3.1-8B & 84.3 & 74.3 & 72.1 & 63.2 & 91.5 & 86.6 & 91.5 & 86.6 & 84.9 & 82.4 \\

% \textbf{EoG} & Llama-3.1-8B & \textbf{92.8} & \textbf{80.7} & \textbf{84.3} & \textbf{73.9} & \textbf{96.1} & \textbf{91.6} & 96.1 & 91.6 & \textbf{85.3} & \textbf{84.3} \\

\bottomrule
\end{tabular}
}
\end{table*}

Beyond reporting mean performance scores, we further validated the reliability of our results through statistical hypothesis testing. Table \ref{tab:ttest_clean} details the t-statistics and p-values comparing EoG with the GCR baseline.  In all datasets, We performed a standard t-test, yielding a p-value $<$ 0.05, which confirms that the performance gains of EoG over GCR are statistically significant and not due to random variance.

\begin{figure}[htbp] 
    \centering
    \includegraphics[width=0.9\linewidth]{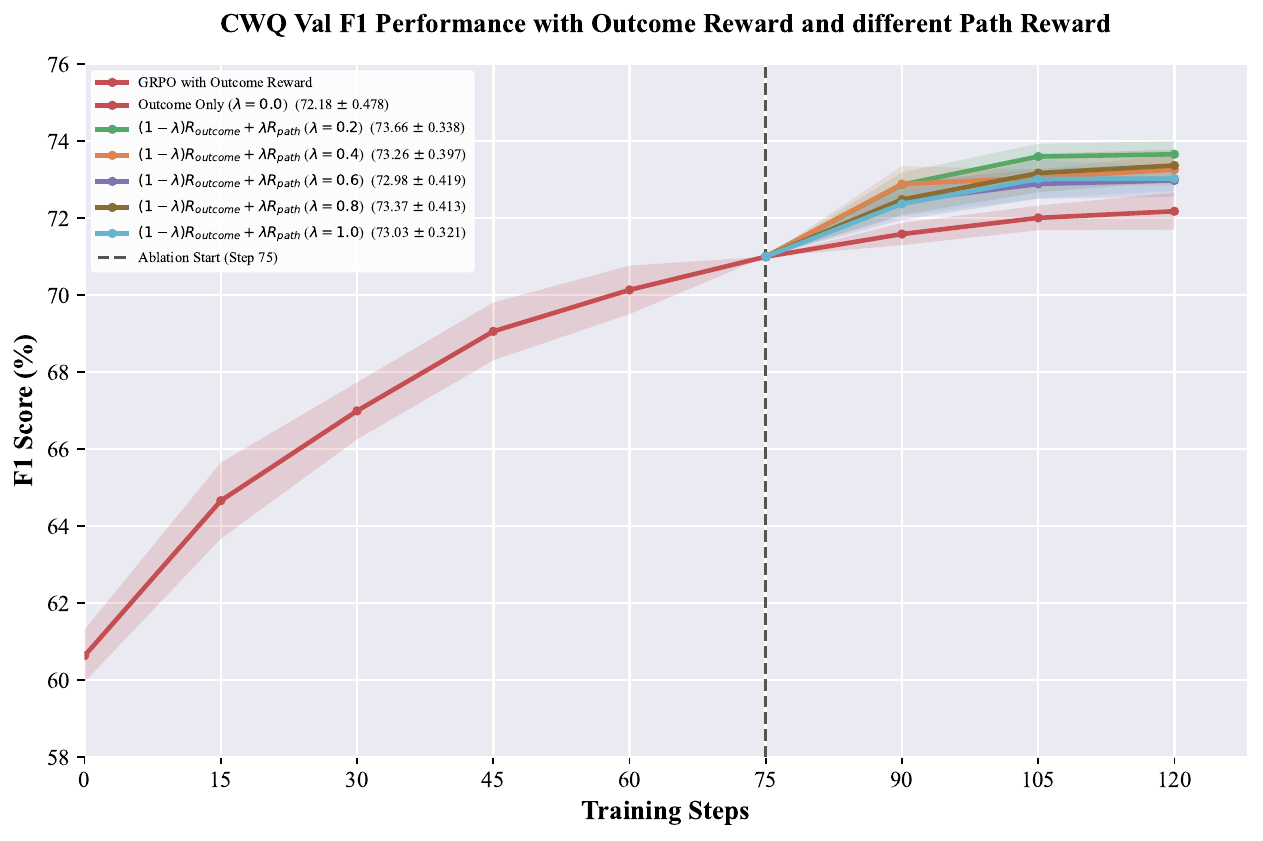}
    \caption{Validation F1 performance on the CWQ dataset with varying Path Reward weights. The total reward is formulated as $R =(1-\lambda)R_{outcome} + \lambda R_{path}$, where $\lambda$ denotes the coefficient shown in the legend. The dashed vertical line marks the start of the second training phase at step 75, where the reward shifts from result-only RL to a mixed result-and-path reward. Solid lines and shaded regions represent the mean and standard deviation across 3 random seeds, respectively. The results indicate that $\lambda=0.2$ achieves the optimal balance, yielding the highest F1 score and robust stability compared to the baseline ($\lambda=0.0$).}
    \label{fig:seed-pdf}
\end{figure}

\begin{table*}[t]
\centering
\caption{Statistical comparison between GCR and our EoG framework using a one-sample t-test ($df=4$). The table reports the baseline score, our model's performance (Mean $\pm$ SD), and the calculated $t$-statistics and $p$-values. Results indicate statistically significant improvements across all metrics.}
\label{tab:ttest_clean}
\setlength{\tabcolsep}{8pt} 
\renewcommand{\arraystretch}{1.1} 
\resizebox{0.8\textwidth}{!}{
\begin{tabular}{llcccc}
\toprule
\multirow{2}{*}{Dataset} & \multirow{2}{*}{Metric} & GCR & EoG (Ours) & \multirow{2}{*}{$t$-value} & \multirow{2}{*}{$p$-value} \\
& & (Baseline) & (Mean $\pm$ SD) & & \\
\midrule
\multirow{2}{*}{WebQSP} 
& Hit@1 & 92.2 & 92.8 $\pm$ 0.28 & 4.79 & 0.009 \\
& F1    & 79.1 & 81.3 $\pm$ 0.33 & 14.91 & $< 0.001$ \\
\addlinespace 
\multirow{2}{*}{CWQ}    
& Hit@1 & 75.8 & 82.6 $\pm$ 0.41 & 37.08 & $< 0.001$ \\
& F1    & 61.7 & 73.9 $\pm$ 0.20 & 136.46 & $< 0.001$ \\
\addlinespace
\multirow{2}{*}{GrailQA}
& Hit@1 & 88.4 & 92.1 $\pm$ 0.30 & 27.57 & $< 0.001$ \\
& F1    & 85.1 & 90.6 $\pm$ 0.25 & 49.20 & $< 0.001$ \\
\addlinespace
\multirow{2}{*}{QALD10} 
& Hit@1 & 57.6 & 70.6 $\pm$ 0.49 & 59.33 & $< 0.001$ \\
& F1    & 49.3 & 61.9 $\pm$ 0.38 & 74.12 & $< 0.001$ \\
\addlinespace
\multirow{2}{*}{2WikiMultihopQA}
& Hit@1 & 84.6 & 85.3 $\pm$ 0.23 & 6.80 & $0.002$ \\
& F1    & 77.5 & 84.3 $\pm$ 0.44 & 34.56 & $< 0.001$ \\
\bottomrule
\end{tabular}
}
\end{table*}
As illustrated in Figure \ref{fig:seed-pdf}, RL typically exhibits high variance during the training phase, as represented by the shaded error bands. It can be observed from Figure \ref{fig:seed-pdf} that the model incorporating the path reward mechanism consistently achieves superior performance compared to the baseline across all three random seeds. The distinct separation between the error bands of the two methods after the 75th step indicates that our proposed framework achieves stable convergence and that the performance gain is statistically significant rather than a result of random variance.

\section{Construction of Ground-Truth Reasoning Paths}
For datasets lacking explicit reasoning paths (e.g., WebQSP, CWQ, GrailQA, and QALD10-en), we constructed ground-truth paths ($r_{g}$) using a two-stage "Search-and-Verify" pipeline. This process bridges the gap in existing benchmarks by deriving reliable reasoning chains from raw Question-Answer (QA) pairs.
\textbf{Path Retrieval via Breadth-First Search (BFS).}
We first employed BFS on the Knowledge Graph to retrieve all potential topological paths connecting the topic entity (identified in the question) to the ground-truth answer entity. This step ensures high recall of potential reasoning chains.
\textbf{Semantic Verification via LLM.}
Since BFS often yields "spurious paths"—paths that are topologically valid but logically unrelated to the question—we utilized an LLM (Gemini-2.5-Flash) as a semantic filter. The LLM was prompted to evaluate whether each retrieved path logically corresponds to the semantic intent of the natural language question. Only paths that passed this semantic verification were retained as $r_{g}$ for calculating $R_{path}$.

Consequently, this approach provides the necessary supervision signals for the second-stage RL training, addressing the lack of $r_g$ in open-source benchmarks.

\section{Detailed Definition of Exploration Metrics}
\label{appendix:exploration_metrics}

To provide a fine-grained analysis of exploration behavior, we parse the 
triples mentioned in the model's reasoning trace (the \texttt{<think>} block) 
on the CWQ test set.

Let $\mathcal{D}$ denote the test set with size $N = |\mathcal{D}|$. 
For the $i$-th question, let $T_{pred}^{(i)}$ be the set of valid unique 
triples extracted from the model's reasoning path, and 
$T_{gold}^{(i)}$ be the set of triples in the ground-truth reasoning path.

\textbf{Exploration Efficiency ($\downarrow$).}
This metric measures the search overhead, representing the average number of triples the model explores to identify one correct reasoning triple. 
A lower value indicates higher efficiency, implying less wasted exploration effort for each valid discovery.
It is calculated as the mean ratio of total extracted triples to the number of correct triples found:

\begin{equation}
    \textbf{Exploration Efficiency} = \frac{1}{N} \sum_{i=1}^{N} 
    \frac{|T_{pred}^{(i)}|}
    {|T_{pred}^{(i)} \cap T_{gold}^{(i)}|}
\end{equation}

\textbf{Coverage of the Reasoning Space ($\uparrow$).}
To define this metric, we first conceptualize the Reasoning Space for a given question as the set of all potential valid paths within the Knowledge Graph that connect the starting entity to the target answer entities. Consequently, this metric measures the reasoning recall, quantifying the average proportion of the ground-truth logical chain (a subset of the reasoning space) that is successfully uncovered by the model's exploration.
A higher value indicates better coverage, meaning the model finds more of the required logical steps.
It is calculated as:
\begin{equation}
    \textbf{Coverage} = \frac{1}{N} \sum_{i=1}^{N} 
    \frac{|T_{pred}^{(i)} \cap T_{gold}^{(i)}|}
    {|T_{gold}^{(i)}|}
\end{equation}

\end{document}